\documentclass[10pt,journal,twoside,web]{ieeecolor}
\usepackage{generic}
\usepackage{cite}
\usepackage{amsmath,amssymb,amsfonts}
\usepackage{algorithmic}
\usepackage{graphicx}
\usepackage{textcomp}
\usepackage{array}
\usepackage{siunitx}
\usepackage{rotating}
\usepackage{amssymb}
\usepackage{colortbl}

\usepackage{booktabs,xcolor,colortbl}
\usepackage{xparse}
\ExplSyntaxOn
\NewDocumentCommand{\longdash}{ O{2} }
 {
  --\prg_replicate:nn { #1 - 1 } { \negthinspace -- }
 }
\ExplSyntaxOff

\newcommand{\ps}{\textcolor[rgb]{0.0,0.0,0.0}} 

\def\BibTeX{{\rm B\kern-.05em{\sc i\kern-.025em b}\kern-.08em
    T\kern-.1667em\lower.7ex\hbox{E}\kern-.125emX}}
\makeatletter
\let\NAT@parse\undefined
\makeatother
\usepackage[pagebackref,breaklinks,colorlinks,citecolor=cyan]{hyperref}

\begin{document}
\title{Automated Radiology Report Generation: \\ A Review of Recent Advances}
\author{Phillip Sloan, Philip Clatworthy, Edwin Simpson, and Majid Mirmehdi
\thanks{P. Sloan was supported by the UK Research and Innovation (UKRI) Centre for Doctoral Training (CDT) in Interactive Artificial Intelligence Award (EP/S022937/1)}
\thanks{P. Sloan is with the Interactive Artificial Intelligence CDT at the School of Computer Science, University of Bristol, UK (e-mail: phillip.sloan@bristol.ac.uk). }
\thanks{P. Clatworthy is with the Bristol Medical School, University of Bristol, UK (e-mail: phil.clatworthy@bristol.ac.uk).}
\thanks{E. Simpson is with the Intelligent Systems Labs, School of Engineering Mathematics and Technology, University of Bristol, UK (email: edwin.simpson@bristol.ac.uk).}
\thanks{M. Mirmehdi is with the School of Computer Science, University of Bristol, UK (email: majid@cs.bris.ac.uk).}}

\maketitle

\begin{abstract}
\ps{Increasing demands on medical imaging departments are taking a toll on the radiologist's ability to deliver timely and accurate reports. Recent technological advances in artificial intelligence have demonstrated great potential for automatic radiology report generation (ARRG), sparking an explosion of research. 
This survey paper conducts a methodological review of contemporary ARRG approaches by way of (i) assessing datasets based on characteristics, such as availability, size, and adoption rate, (ii) examining deep learning training methods, such as contrastive learning and reinforcement learning, (iii) exploring state-of-the-art model architectures, including variations of CNN and transformer models, (iv) outlining techniques integrating clinical knowledge through multimodal inputs and knowledge graphs, and (v) scrutinising current model evaluation techniques, including commonly applied NLP metrics and qualitative clinical reviews.    
Furthermore, the quantitative results of the reviewed models are analysed, where the top performing models are examined to seek further insights. Finally, potential new directions are highlighted, with the adoption of additional datasets from other radiological modalities and improved evaluation methods predicted as important areas of future development.}

\end{abstract}

\section{Introduction}
\label{sec:introduction}

\begin{figure}[!ht]
\centerline{\includegraphics[width=\columnwidth]{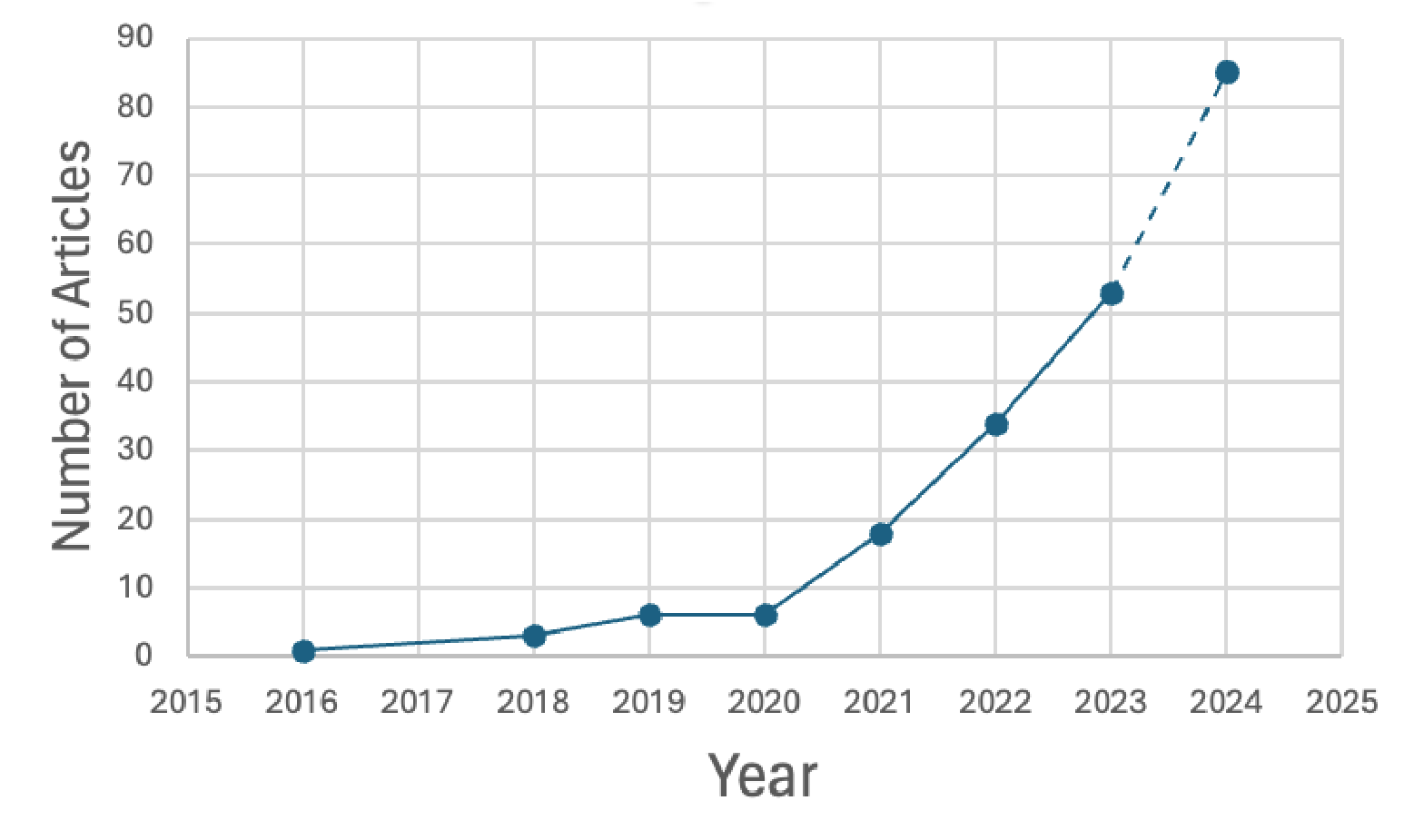}}
\caption{Illustration of the rapid growth of published papers within the ARRG domain. The dashed line represents a forecast of anticipated future publications.}
\label{fig:rapidgrowth}
\end{figure}

\IEEEPARstart{R}{adiology} is critical to modern healthcare and is the main diagnostic tool used by clinicians for monitoring treatment and predicting outcomes. Radiology includes various imaging modalities, and after each examination has been performed, a radiological report is produced to guide patient management. The diagnostic accuracy of such reports is critical in ensuring optimal care, however it has been reported that there is a discrepancy rate of 3--5\% \cite{Brady2016} while expressions of uncertainty have been found in 35\% of reports \cite{Mabotuwana2018}. This can be compounded by a shortage of radiologists in some regions, for example  97\% of imaging departments in the UK report that they can not keep up with demand \cite{Abi}. Such understaffing can cause reporting delays leading to clinicians making critical decisions without the benefit of a radiological opinion and potentially reaching
a different conclusion compared to that of an interpreting radiologist \cite{Petinaux2011}.

Healthcare represents an important application of deep learning research and with \ps{r}adiology producing 90\% of healthcare data \cite{Zhou2021}, it is an attractive area for deep learning researchers. Methods from computer vision (CV) \cite{Cai2020} and natural language processing (NLP) \cite{Zhang2020, Chng2023} are currently being adopted within the healthcare domain to improve the affordability and standards of care. One rapidly developing healthcare application of deep learning is automated radiology report generation (ARRG), a vision-language task that has similarities to the broader area of image captioning \cite{Wang2022}.
With its ability to augment radiologists' capabilities, ARRG has significant clinical value 
and could alleviate time pressures by reporting relatively simple examinations. It could also be used as a patient specific draft template and for an inexperienced radiologist it could act as a second reader \cite{RCR} by automatically highlighting any potential abnormalities \cite{Yang2022}.

\begin{figure*}[ht]
\centerline{\includegraphics[width=\textwidth]{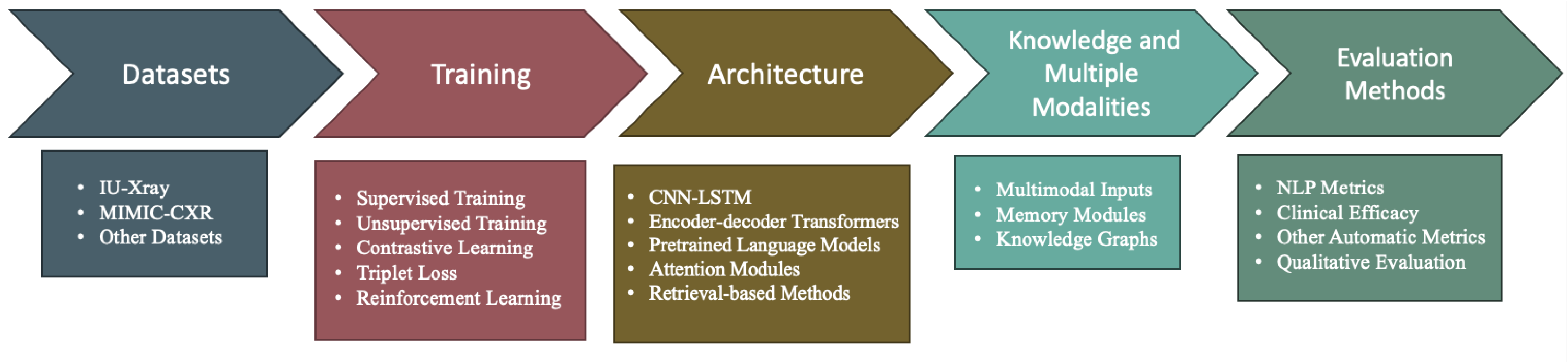}}
\caption{The road map of this paper -- First relevant ARRG datasets are discussed, before we visit recent model training  approaches. Next, the range and extent of deep learning architectures deployed or developed are discussed, followed by the ``who and how'' of works that have used knowledge and multiple modalities. Finally, we consider assessment and evaluation methods employed in, or specifically crafted for, ARRG.}
\label{fig:roadmap}
\end{figure*}

There have been prior surveys of the ARRG domain \cite{Monshi2020, Kaur2022, Liao2023} and of the automated medical report generation domain \cite{Allaouzi2018, Pavlopoulos2019, Messina2022} (see Table~\ref{tab:datasets} for summary).
While these earlier papers offered an analysis of the initial growth of ARRG, there has since been a surge of papers released over the past 2-3 years as can be observed in Figure~\ref{fig:rapidgrowth}. \ps{We anticipate this trend to continue, driven by the rising number of researchers in this field who seek to harness technological advances in various areas, such as language modelling \cite{llama, falcon}, knowledge graphs \cite{Li2023, Zhao2022}, and contrastive learning \cite{Hou2023b, Jeong2023, WuX2023, Ye2023}. Moreover, in early 2024, a new 3D chest computed tomography (CT) dataset was made available \cite{Hamamci2024}. Although no research has yet been published using this dataset, it is anticipated that researchers will seek to leverage the additional information it offers, potentially resulting in an even greater influx of papers in ARRG.}


\ps{Our review seeks to cover the abundance of recent works since the latest review papers of Kaur et al. \cite{Kaur2022} and Messina et al. \cite{Messina2022}.}
In particular, Kaur et al. \cite{Kaur2022} provided a comprehensive survey of the ARRG domain specific to chest radiographs. They classified research into four categories including encoder–decoder architectures, attention mechanisms, reinforcement learning, and graphs. While we also cover these areas in our review, there has been an influx of research into new areas, e.g. training methods such as curriculum learning \cite{Liu2021} and contrastive learning \cite{Hou2023b} and the development of qualitative evaluation methods such as assessment by clinical experts \cite{Yan2021}. As their review includes papers up to and including 2021, we seek to complement their survey by reviewing these new avenues of research only from the last few years. 

Liao et al \cite{Liao2023}'s survey spanned publications released by 2021 in relation to datasets, training, architecture, interpretability and evaluation metrics. Since the publication of their review, there has been a deluge in research investigating injecting semantic knowledge into the model through the use of knowledge graphs, such as RadGraph \cite{Radgraph}. Our review builds on their review, but also evaluates approaches injected with semantic knowledge, e.g. \cite{Wang2022, WangZ2022, Zhang2022} 
and ARRG contributions with a multimodal input, where the input contained not only images, but other information such as clinical indications \cite{Nguyen2021, Shang2022, DallaSerra2022, DallaSerra2023, Mondal2023} or temporal features \cite{Nishino2022, Bannur2023, DallaSerra2023, Zhu2023}.

The main contribution of this work is a comprehensive review of the most recent literature on automated radiology report generation since 2020,
examining the research by way of five different categories: datasets, training, architecture, utilising knowledge and multiple modalities, 
and evaluation methods {(see road map in Figure~\ref{fig:roadmap})}.  
We offer carefully curated figures and tables to distinguish papers based on the alignment of their research contributions with these five
classifications. 
The articles reviewed in this study are predominantly sourced from high impact journals and conferences, such as Neural Information Processing Systems (NeurIPS), Computer Vision Pattern Recognition (CVPR), Empirical Methods in Natural Language Processing (EMNLP), Institute of Electrical and Electronics Engineers(IEEE) and Medical Image Computing and Computer Assisted Interventions (MICCAI). Occasionally, when significant and helpful to the narrative, we include work prior to 2020.

The rest of the paper is organised as follows. 
Section~\ref{sec:background} provides a background to the ARRG task. 
Section~\ref{sec:datasets} outlines the datasets used for ARRG. 
\ps{Section~\ref{subsec:training} reviews articles which seek to improve different aspects of model training.}
\ps{Section~\ref{subsec:transformers} discusses contributions which develop  model architecture.}
\ps{Section~\ref{subsec:semantic} explores research that utilises domain knowledge or multiple modalities within their models.}
\ps{Section~\ref{subsec:eval} considers the various evaluation methods used within the ARRG domain.}
Section~\ref{sec:evaluation} discusses the ARRG landscape. We contrast the reported model performances on commonly employed NLP metrics.
Section~\ref{sec:conclusion} concludes the paper, discussing potential future directions in the field and outlining the potential limitations of this review.

\begin{table}
\scriptsize
    \centering
    \begin{tabular}{lccrr} \hline 
         \textbf{~~~~~Author}&  \textbf{Year}&  \textbf{Coverage}&  \textbf{\# Cited}&  \textbf{\# ARRG}\\ \hline 
         Allaouzi et al. \cite{Allaouzi2018} &  2018&  2015-2018&  50&   10\\ \hline 
         \rowcolor[gray]{0.9}
         Pavlopoulos et al. \cite{Pavlopoulos2019}&  2019&  2015-2019&  65&   7\\ \hline 
         Monshi et al. \cite{Monshi2020}&  2020&  2015-2019&  105&   14\\ \hline 
         \rowcolor[gray]{0.9}
         Kaur et al. \cite{Kaur2022}&  2022&  2015-2021&  83&   19\\ \hline 
         Messina et al. \cite{Messina2022}&  2022&  2016-2021&  167&   40\\ \hline 
         \rowcolor[gray]{0.9}
         Liao et al. \cite{Liao2023}&  2023&  2016-2021&  154&   41\\ \hline
         Our Review& 2023& 2020-2023& \ps{155}&\ps{84}\\\hline
    \end{tabular}
    \caption{Recent ARRG review articles and their coverage including total number of citations and the total number of articles related to ARRG or medical report generation.}
    \label{tab:datasets}
\end{table}

\section{Background}
\label{sec:background}
The purpose of a radiology report is to provide an accurate interpretation of an imaging study relating to the findings that have been observed within the study, irrespective of whether they are anticipated or unexpected. Reports should also be specific in answering the clinical question asked by the referring clinician who then uses information gleaned from the report to provide appropriate care for the affected patient. While there is no universally agreed structure to radiology reports, the European Society of Radiology \ps{(ESR)} \cite{ESR2011} have outlined good practice guidelines for the standardised structure of a radiological report:

\begin{itemize}
    \item \textbf{Clinical referral --} A brief summary of the indications and reasons for the referral.  This often includes the clinical question asked by the referring clinician. 
    \item \textbf{Technique --} A description of what investigation was performed and which techniques or sequences were utilised.
    \item \textbf{Findings --} A comprehensive description of all observations within the image, starting with the most relevant to the request. The findings should note any incidental findings not related to the indications.
    \item \textbf{Conclusion/Impression --} A concise diagnosis of the examination or a ranked list of potential differential diagnoses to help guide patient treatment.
\end{itemize}

\begin{figure}[!ht]
\centerline{\includegraphics[width=\columnwidth]{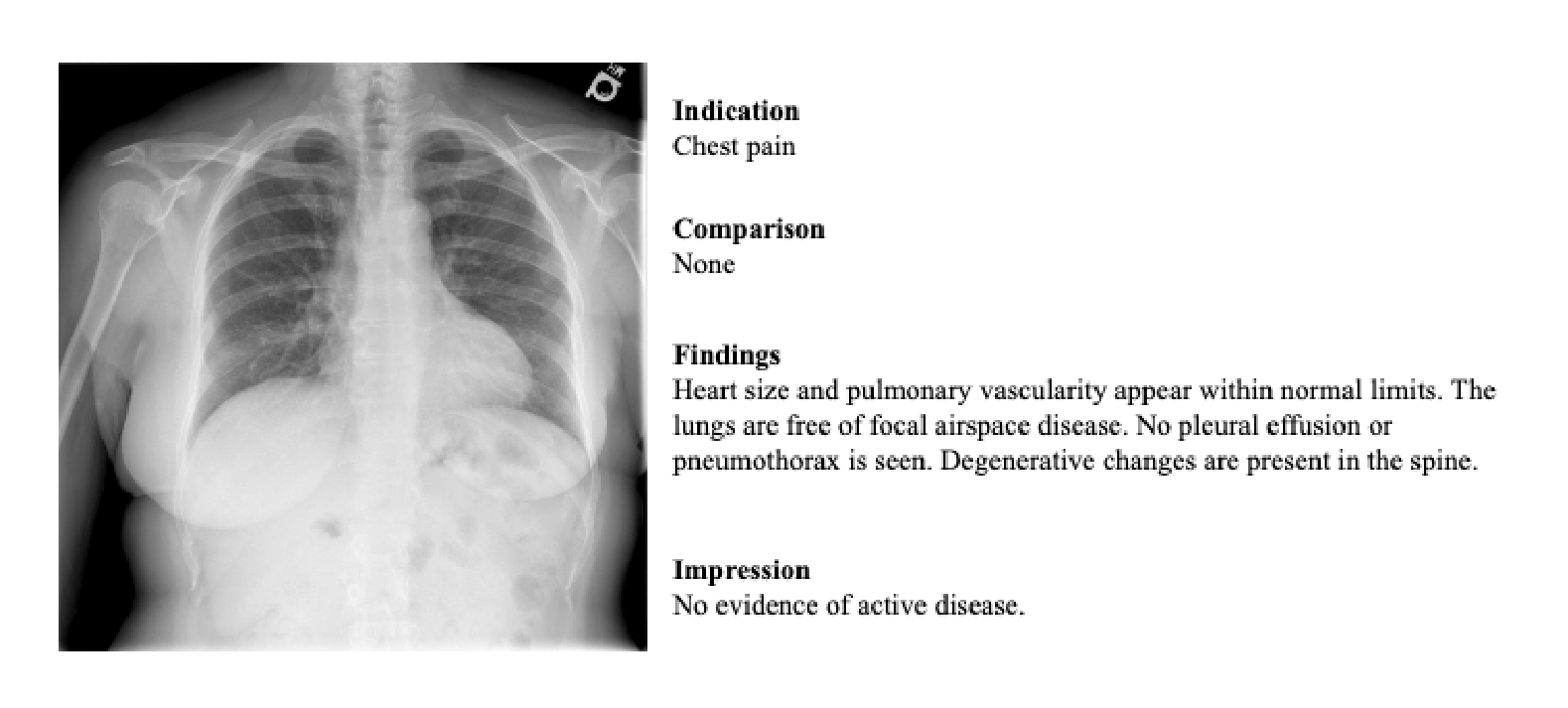}}
\caption{An example x-ray and report pair from Indiana University X-ray dataset \cite{IU-Xray}. ARRG models seek to generate the Findings and Impression sections.}
\label{fig:iuxrayexample}
\end{figure}

The ARRG task takes one or more radiographic images as input and attempts to produce a sequence of text that closely resembles the report created by a radiologist.

Deep learning approaches for ARRG have generally adopted generative models, with an encoder-decoder architecture being the most commonly used. Within this architecture the encoder is used to extract meaningful features from the input radiographic image, and subsequently, the decoder used the extracted features to generate the radiology report. The majority of reviewed research trains in a supervised fashion where the associated report is used as a ground truth for training and evaluation. An example of a training sample can be seen in Figure~\ref{fig:iuxrayexample}.

Early ARRG research coincided with the 2016 release of the Indiana University X-ray Dataset \cite{IU-Xray}, the first dataset that contained both radiology images and their associated radiology reports. Shin et al. \cite{Shin2016} used this dataset to develop a method to automatically annotate medical images using a convolutional neural network (CNN) for their encoder and a recurrent neural network (RNN) for their decoder. In 2018, research into ARRG started gaining popularity \cite{Xue2018, Jing2018}, however the generated reports from these CNN-RNN methodologies are overly rigid \cite{Wang2022} and often repeat phrases from the training set at inappropriate times, leading to summaries that appear less human-like. The development of the transformer architecture \cite{Vaswani} has created new avenues to be explored and stimulated significant amounts of research in the ARRG field.

ARRG can be considered to have its roots in image captioning. From the CV perspective, unlike natural images such as images of cats and dogs, \ps{radiological images closely resemble each other with differences between two radiographic images being quite nuanced \cite{Wang2022}}. A pathology could occupy only a limited section of an image, thereby increasing the burden on the encoder. These subtle changes have a high impact on the generated report \ps{and if missed, will turn an abnormal case into a normal one.} 
From an NLP perspective, the generated text is longer than a traditional caption. In addition, differences in reports from patient to patient are fine-grained, meaning there are a lot of similar sentences describing the normal appearances, while the sentences related to observed abnormalities could be overlooked \cite{Wang2022}. 

\section{Datasets}
\label{sec:datasets}
\begin{table*}
\scriptsize
    \centering
    \begin{tabular}{lccclrrr cr} \hline 
         \textbf{~~~~~Name}&\textbf{Year}&\textbf{Availability} &\textbf{Modality}  &\textbf{~~~~Views}&\textbf{Patients} &\textbf{E/R}&\textbf{Images}&\textbf{Language} &\textbf{Adoption~~~}\\  \hline 
         IU-Xray \cite{IU-Xray}&2016& Public &CXR  &\ps{Frontal/Lateral} &\ps{3,955} &\ps{3,955}&7,470 &English &\ps{50}\\ \hline 
         \rowcolor[gray]{0.9}
         MIMIC-CXR \cite{MIMIC}&2019& \ps{Restricted} &CXR  &\ps{Frontal/Lateral} &\ps{65,379} &\ps{227,827}&377,110 &English &\ps{45}\\\hline 
         MIMIC-ABM \cite{MIMIC-Abm}& 2020& \ps{Restricted}& CXR  &\ps{Frontal/Lateral} & \longdash~~~~&\longdash~~~~&\ps{38,551}&English &\ps{0}\\\hline 
         \rowcolor[gray]{0.9}
        
          \ps{Chest ImaGenome} \cite{Wu2021}& 2021& \ps{Restricted}& CXR  &\ps{Frontal} &\longdash~~~ &\longdash~~~~&\ps{242,072}&English &\ps{1}\\\hline
         
         \ps{CXR-PRO} \cite{Ramesh2022}& 2022& \ps{Restricted}& CXR  &\ps{Frontal/Lateral} &\ps{65,379} &\longdash~~~~&374,139&English &\ps{1}\\\hline

         \ps{MS-CXR} \cite{Boecking2022} & 2022& \ps{Restricted}& CXR  &\ps{Frontal} &\ps{851} &\longdash~~~~&1047&English &\ps{1}\\\hline

         \ps{MS-CXR-T} \cite{Bannur2023} & 2023& \ps{Restricted}& CXR  &\ps{Frontal} &\ps{800} &\longdash~~~~&1326&English &\ps{1}\\\hline

         \rowcolor[gray]{0.9}
         PadChest \cite{PadChest}&2019& \ps{Restricted} &CXR  &\ps{Frontal/Lateral} &\ps{67,625} &\ps{109,931}&160,868 &Spanish &\ps{0}\\ \hline 
         
         CH-Xray \cite{Zhao2022}&2022& Private &CXR  &\ps{Frontal} & \longdash~~~~&\ps{11,049}&11,049 &Chinese &\ps{3}\\ \hline
         \rowcolor[gray]{0.9}
         CX-CXR \cite{WangF2022}&2018& \ps{Restricted} &CXR  &\ps{Frontal/Lateral} &\ps{33,236} &\ps{33,236}&45,598 &Chinese &\ps{2}\\ \hline          
         COV-CTR \cite{Li2023}&2022& Public& CT  &\ps{Axial} &\longdash~~~ &\ps{728}&728 & English &\ps{1}\\\hline
         \rowcolor[gray]{0.9}
         JLiverCT \cite{Nishino2022}&2022& Private& CT  &\ps{Axial} & \longdash~~~~&\ps{1,083}&1,083 &Japanese &\ps{1}\\\hline
         \ps{CT-RATE} \cite{Hamamci2024}&\ps{2024}& \ps{Public}& \ps{CT}  &\ps{Axial} &\ps{21,304} &\ps{25,692}&\ps{50,188} &\ps{English} &\ps{1}\\\hline
    \end{tabular}
    \caption{Datasets used in the Literature. Restricted means an application process is required to gain access to the associated dataset. A `\longdash'~indicates that the corresponding value was not reported for this dataset. E/R represents the number of Examinations and Reports. The adoption field quantities how many of the reviewed articles use a dataset. }
    \label{tbl:datasets}
\end{table*}

Ethical and confidentiality issues surrounding the public release of medical data \cite{Fernandez2023} means medical datasets with associated radiology reports are scarce. 
In this section, after considering two publicly available datasets that are used most frequently by researchers in ARRG, we shall briefly mention a number of other datasets that have been harnessed by ARRG researchers to improve their models. 
A list of the most significant datasets encountered in our search of the existing literature, where actual reports beyond simple patient metadata are available, is available in Table~\ref{tbl:datasets}. 

\subsection{Indiana University X-ray Dataset (IU-Xray)}
\label{subsec:iuxray}
The Indiana University X-ray Dataset dataset (IU-Xray) \cite{IU-Xray}, also known as the OpenI Dataset, was published in 2016. It is publicly available and contains $7,470$ images and $3,955$ radiology reports. Each report is comprised of four sections: indication, comparison, findings and impression, and thus can be considered to be roughly aligned with ESR guidelines. IU-Xray has also been labelled with Medical Text Indexer (MTI) tags which are keywords generated from the diagnostic sentences within the report. While there is no official split, some more recent researchers \cite{Chen2021, LiuF2021, Huang2023, Wang2023} have adopted the split provided by Chen et al. \cite{Chen2020} for reproducibility. 

\ps{When reviewing Table~\ref{tbl:datasets}, IU-Xray emerges as the most prevalent dataset, with $50$ of the reviewed articles using this dataset for training and evaluation. This can be attributed to its easy accessibility and relatively small size which allows quick prototyping and training. It's relatively small size has however been noted as a potential limitation, with Chen et al. \cite{Chen2020} stating that its size can result in significant variance of results. Additionally, IU-Xray is composed of only one examination per patient, meaning there is no longitudinal dimension for researchers to leverage. Moreover, the data collection exclusively included outpatient examinations, thereby potentially impeding the accuracy and generalisability of any trained models due to the absence of certain projections and artefacts, such as portable x-rays and central lines, which are out of distribution.}

\ps{While IU-Xray is a very popular dataset within the ARRG domain, it has been noted that researchers sometimes employ the dataset as a secondary testing set \cite{Chen2020}.}

\subsection{Medical Information Mart for Intensive Care CXR (MIMIC-CXR)}
\label{subsec:mimic}
The Medical Information Mart for Intensive Care Chest X-ray (MIMIC-CXR) \cite{MIMIC} stands as the largest ARRG dataset to date, comprising $377,110$ radiographic images and $227,835$ radiology reports collected from $65,379$ individuals. These records span the years 2011-2016 and originate from the Beth Israel Deaconess Medical Center Emergency Department in Boston, MA. In addition to the images and reports, the dataset also contains medical subject headings (MeSH), which are labels that are automatically generated, annotating 13 important medical tags for each sample. The original release of MIMIC-CXR was in Digital Imaging and Communications in Medicine (DICOM) format, however the dataset has since been converted into the \ps{JPEG} image format. MIMIC-CXR has an official train/test/validation split, which has easily enabled comparisons between different ARRG models. \ps{Researchers who wish to access MIMIC-CXR are required to be credentialed on PhysioNet \footnote{https://physionet.org/} and to sign a data use agreement.}

\ps{When examining the datasets listed in Table~\ref{tbl:datasets}, it becomes apparent from the adoption column that MIMIC-CXR is one of the main choices among researchers in the field of ARRG. It also stands out as the largest available dataset for ARRG which is advantageous because larger datasets often lead to reduced model variance \cite{Ying2019}. The ratio of patients to examinations suggests that each patient has on average $3.5$ studies within the dataset, which highlights the temporal aspect of MIMIC-CXR, presenting researchers with significant potential for exploitation. MIMIC-CXR is also part of the larger MIMIC database which contains various other related datasets that could potentially be used in conjunction with with MIMIC-CXR.}

\subsection{MIMIC Derivatives}
\ps{Due to its rich data and popularity among researchers, several derivative datasets have been developed from MIMIC-CXR. These datasets are located on PhysioNet and similarly to MIMIC-CXR, require researchers to be credentialed before they can gain access.} Ni et al. \cite{MIMIC-Abm} developed a subset called MIMIC-ABM which is constructed from samples of MIMIC-CXR \ps{that have at least one abnormal finding in the associated report.} \ps{Ramesh et al \cite{Ramesh2022} implemented another derivative called CXR-PRO that omits from reports any reference to prior radiological studies through the use of a BERT-based token classifier.}

\ps{Chest ImaGenome was constructed by Wu et al. \cite{Wu2021}, which contains automatically constructed scene graphs for the MIMIC-CXR images. Each scene graph describes one frontal chest X-ray image and contains bounding box coordinates for $29$ unique anatomical regions in the chest, as well as sentences describing each region if they exist in the corresponding radiology report. The dataset is split according to the official MIMIC split. Boecking et al. \cite{Boecking2022} developed a dataset called MS-CXR which contains $1047$ images with a total of $1162$ image–sentence pairs of bounding boxes and corresponding phrases, collected across eight different cardiopulmonary radiological findings, with an approximately equal number of image-sentence pairs for each finding.}

\ps{Taking advantage of the longitudinal nature of MIMIC, Bannur et al \cite{Bannur2023} released MS-CXR-T, a multi-modal benchmark dataset for evaluating biomedical vision-language models on two distinct temporal tasks in radiology: image classification and sentence similarity. The image classification task provides labels across five findings that can be considered to be in one of three states: improving, stable or worsening. The sentence similarity task contains $361$ pairs of sentences 
that are either semantically equivalent or contradictions. The datasets aim was to provide an opportunity to evaluate models on temporal tasks, which has not previously been available within the ARRG domain.}


\subsection{Other Datasets} \label{sec:otherdatasets}
Aurelia et al. \cite{PadChest} compiled PadChest which is a dataset of 160,868 images and 109,931 Spanish reports from the San Juan Hospital in Spain. Although PadChest contains public reports, researchers have not embraced it, and the reason for this lack of adoption remains unclear. Typically, having datasets from various languages is deemed advantageous for assessing whether a method generalises across languages. To gain access to PadChest, a researcher will need to apply to the principle researchers through an application form. Li et al. \cite{Li2023} publicly released COV-CTR, a COVID-19 CT Report dataset containing 728 samples, which contains reports both in English and Chinese. CX-CHR\cite{WangF2022} is a private collection of chest x-ray images with corresponding Chinese reports consisting of 33,236 samples. While not publicly available, the paper states that researchers can apply for academic usage after signing a confidentiality agreement.

\ps{CT-RATE \cite{Hamamci2024} is a recently published, 3D CT Chest dataset, containing 25,692 non-contrast chest CT volumes. It has been expanded to 50,188 volumes through various reconstructions and encompasses data from 21,304 unique patients 
, accompanied by the corresponding English radiology text reports. CT-RATE provides an important step forward for the ARRG domain as it is the first large scale dataset in a radiology modality other than 2D chest x-rays. Similar to IU-Xray \cite{IU-Xray} and MIMIC-CXR \cite{MIMIC}, the reports contain clinical indications, findings and impressions. The dataset is publicly available through a HuggingFace dataset repository \footnote{https://huggingface.co/datasets}.}

There are very few other datasets used in ARRG research often limited by their restricted (or complete lack of) accessibility.
CH-Xray \cite{Zhao2022} is a private dataset with Chinese language reports containing 11,049 samples. The private JLiverCT dataset \cite{Nishino2022} contains 1,083 samples, with the reports written in Japanese. {Liu et al. \cite{Liu2023Dataset} released a public dataset of patients experiencing acute and early sub-acute stroke, comprising of 2,888 multi-sequence clinical MRIs along with metadata (age, sex and other clinical factors). Subsequently, Liu et al. \cite{Liu2023ARRG}  generated reports for this dataset, but these are not accessible to the public at present which inhibits the wider ARRG community from adopting the dataset.}

There are some datasets that do not offer free-text reports, but have been used to help augment a model's ability to handle ARRG. The National Institutes of Health Chest X-Ray Dataset (NIH-CXR) \cite{NIH-Xray} is composed of 112,120 images from 30,805 examinations, with labels text-mined from radiology reports, initially containing 8 labels and improved to 14 on a later release. CheXpert \cite{CheXpert} consists of 224,316 images obtained from the Stanford University Medical Center between 2002--2017 and also uses an NLP method to mine 14 thoracic pathological labels. 



\section{Training}
\label{subsec:training}
Training a computational model to learn how to make decisions and generate knowledge on its input data is a key stage of building such a model. We now categorise, loosely under specific topics, various training strategies that were originally developed for other applications, but have been either adopted or developed further for ARRG.


\subsection{Supervised Learning}
Supervised learning is defined by training a model using datasets containing ground truth labels. In ARRG, the label is the radiology report which is the target output the model is trained to predict. For ARRG datasets, 
some training samples are more challenging than others as the radiographic image could contain a subtle pathology or the radiology report could be particularly long. Curriculum learning is a training strategy where, instead of randomly shuffling the entire dataset, the model is exposed to increasingly complex examples over time. This method of training a model is thought to closely resemble how a radiologist learns throughout their career, enabling a clinician to gradually become more comfortable with more complex cases. The novelty of curriculum learning techniques relates to how a researcher measures how challenging an example case is. 

Liu et al. \cite{Liu2021} developed a competence-based multimodal curriculum learning strategy which evaluated the complexity of each sample in the dataset using heuristic metrics to quantify the visual and textual difficulty. To measure the degree of visual difficulty, they implemented a ResNet-50 architecture, fine-tuned on the CheXpert dataset. Their model extracted the normal image embeddings of all normal training images from ResNet-50's last average pooling layer. Then, given an input image, an image embedding was obtained from their  model which was compared against the normal images 
using average cosine similarity to provide a heuristic measure of the visual complexity of that training instance. For the textual difficulty, they utilised a count of the number of sentences describing abnormal findings within the report to establish how arduous it is to generate. Their method considered sentences without the keywords ``no'', ``normal'', ``clear'' or ``stable'' as sentences describing abnormal findings. 

\begin{figure*}[ht]
\centerline{\includegraphics[width=.9\textwidth]{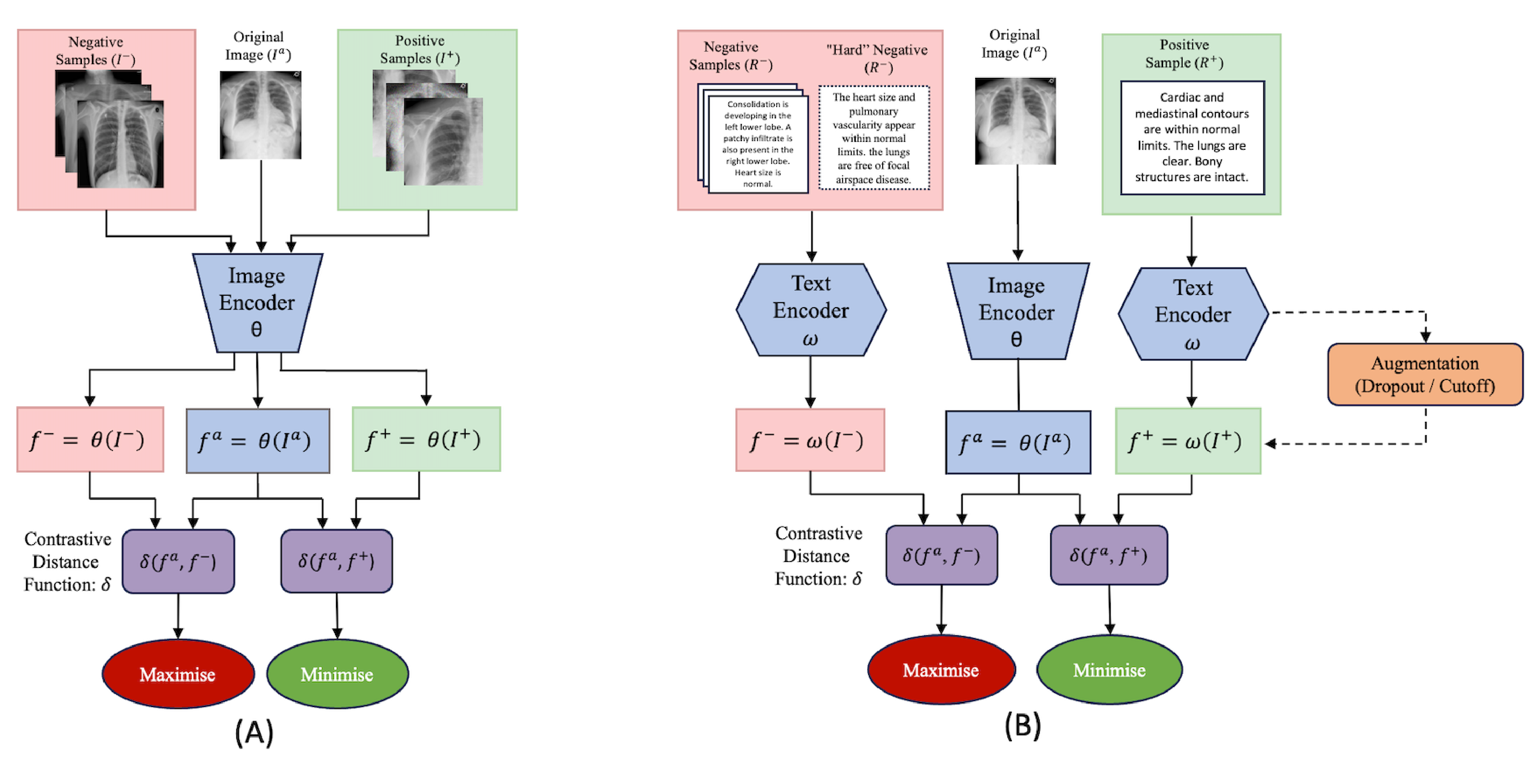}}
\caption{An illustration of (A) contrastive learning and (B) contrastive language-image pre-training (CLIP) methods used within ARRG. (A) demonstrates SimCLR \cite{SimCLR} as implemented for example by \cite{Hou2023b, Tanwani2022}. In this framework, image augmentations such as cropping and Gaussian noise were used to create additional positive samples while other radiographic images were considered as negative samples. (B) demonstrates CLIP, which was implemented by works such as \cite{Endo2021, Leonardi2022}. The dotted lines present novel pathways from ARRG research contributions: ``hard'' negatives were used by  \cite{Yan2021, Jeong2023} to enable more robust representations to be learnt and \cite{WuX2023} developed extra positive samples at the embedding level through the use of dropout and cutoff.}
\label{fig:training}
\end{figure*}

A multi-criteria supervised approach to training was proposed by Wang et al. \cite{Wang2022}, with auxiliary objectives embedded into their training strategy. An image-text matching objective is used to better correlate image and text features while an image classification objective is also developed to improve the feature extraction capabilities of their model. For the classification objective, the MeSH labels provided with the MIMIC-CXR dataset are adopted as ground truths. Li et al. \cite{Li2023} also implemented auxiliary objectives within their model, naming the process auxiliary signal‐guided learning. Their method implemented internal auxiliary signals where their model generated segmentation masks of abnormal regions which are used to crop the radiographic images. These cropped images serve as a secondary input to the model which enabled the model to attend to abnormal areas in more detail. Their model also employed external auxiliary signals to correctly phrase and learn medical knowledge. The medical knowledge is mined from an unnamed, medical website which provided medical knowledge on the symptoms, manifestations and other information regarding thoracic diseases. This medical text is divided into tokens, which are embedded before being encoded with a single-layer gated-recurrent unit. 

\subsection{Unsupervised Learning}
Unsupervised learning trains models using unlabeled datasets, seeking to find hidden patterns and insights from the given data. During our literature search, we encountered 
less than a handful significant ARRG research contributions concerning unsupervised learning for the main ARRG task which suggests the domain is either under-explored or self-supervised approaches have already extinguished the flame for such approaches.

Liu et al. \cite{liu2021autoencoding} leveraged a pre-constructured knowledge graph as a shared embedding space to bridge the visual and textual domains with their knowledge-driven encoder adopting uncoupled image-report pairs in IU-Xray and MIMIC-CXR as queries, projecting them to corresponding coordinates in the latent space. These embedding positions within the latent space are used to measure the relationship between images and reports. 
A variational autoencoder-based unsupervised clustering method was proposed by Yan et al. \cite{Yan2022} to glean additional knowledge 
from input radiographic images. The input images go through a ResNet-101 feature extractor, and then their refinement is carried out by a Transformer encoder. Subsequently, clustering is applied to the refined patches, effectively capturing and representing the additional knowledge. Specifically, additional knowledge is represented by an additive Gaussian distribution which is learned in an unsupervised manner. This prior knowledge is then used as an additional input to their decoder to generate medically fluent reports. Li et al \cite{Li2022} also utilised clustering to obtain potential knowledge from reports, adopting HDBSCAN to identify clusters of similar knowledge. {Rather than randomly initialising their memory matrix module, Wang et al. \cite{WangJ2022} employed K-means clustering on visual and textual features extracted from the training set of the MIMIC-CXR dataset to calculate the initial starting values for their matrix.} 

 

\subsection{Contrastive Learning}
A common form of training implemented in the ARRG domain is contrastive learning \cite{Hou2023b, Leonardi2022, Jeong2023, WangJ2022, You2022, Wu2023, Li2023b, Xu2023} \ps{which focuses on learning representations by comparing similar and dissimilar instances within the data itself.} By contrasting positive and negative pairs of example data, it generates features for similar pairs closer together in a feature embedding space, while dissimilar pairs are registered farther apart. \ps{Depending on how the positive/negative pairings are generated, contrastive learning can be defined as either supervised or unsupervised training, often being classified as self-supervised training due to the lack of explicit labels.}

Contrastive learning approaches have become popular due to the scarcity of data within the medical deep learning domain with works differing in how they select positive and negative samples. Some works have focused on contrasting radiographic images \cite{Hou2023b, Jeong2023, WangJ2022} while others implemented the Contrastive Language–Image Pre-training (CLIP) approach \cite{CLIP}, where both radiographic images and reports are used in some fashion \cite{Yan2021, Endo2021, Tanwani2022, WuX2023, Leonardi2022, Nazarov2022, Ye2023}. An illustration of contrastive learning and CLIP can be seen in Figure~\ref{fig:training}.



Yan et al. \cite{Yan2021} developed a weakly supervised contrastive learning method, comparing the images and radiology reports against each other. The novelty of their approach was through the identification of ``hard'' negative samples which are cases that are close semantically to the report of the corresponding radiographic image. K-means clustering is used to select these ``hard'' samples which originate from the same cluster as the reference report. \ps{Li et al.'s \cite{Li2023b} proposed framework also identified ``hard'' negative samples during training, using cosine similarity to locate the nearest samples in the embedding space.}

Leonardi et al. \cite{Leonardi2022} utilised contrastive learning on radiographic images and reports as pairs. An image and its associated report are used as positive pairs while an image and an alternative report is considered a negative pair. The approach implemented several methods for image or textual augmentations to generate extra positive samples from the reference image/report pair.

Tanwani et al.\cite{Tanwani2022} incorporated SimCLR in their proposed network RepsNet, a visual question answering model which is also shown to be capable at ARRG. Image augmentations such as colour jittering, normalisation and random erasing are used during their approach to generate new positive samples. Their model 
used a bilinear attention network to fuse image and question/text features together after which bidirectional contrastive learning \cite{SimCLR} is adopted, pulling together a given image-text pair while pushing away observations that correspond to a different image-text pairs. Hou et al. \cite{Hou2023b} also drew inspiration from the SimCLR framework \cite{SimCLR} and generated positive samples through augmentation of the radiographic images with transforms such as rotation and Gaussian blurring. These augmented samples, along with any samples from the same patient, are considered positive samples while all other samples are considered negative samples. Wu et al. \cite{WuX2023} generated extra samples through augmenting report sentences at the embedding level using methods such as dropout and cutoff. Augmented sentences are considered positive instances while other sentences within the same batch are considered negative instances.

\begin{figure*}[ht!]
\centerline{\includegraphics[width=.9\textwidth]{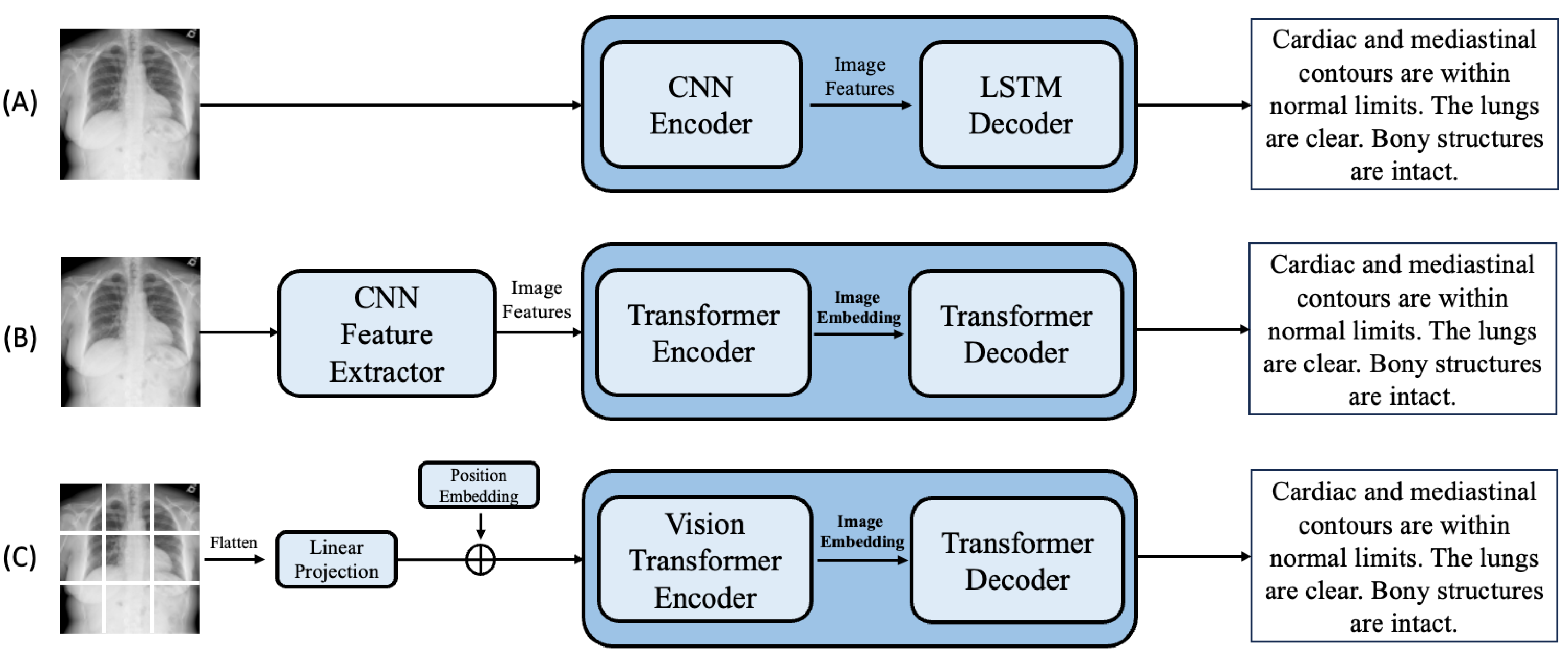}}
\caption{An overview of the three main encoder-decoder architectures used within the ARRG domain: (A) demonstrates a typical CNN-LSTM network as utilised for example by \cite{ZhangY2020, Liu2021b, WangZ2021, WangS2022, Nishino2022, Gajbhiye2022, WangF2022, Kaur2023}, (B) illustrates the most popular architecture in use today, such as in \cite{Chen2020, Hou2023, Wu2023, Huang2023} while (C) embodies the pure transformer architecture used for example by \cite{Wang2022, Shang2022, Mohsan2023}.}
\label{fig:architectures}
\end{figure*}

The CheXpert labeller \cite{CheXpert} is an NLP tool that extracts observations from the reports which have been used in various contrastive learning approaches to help distinguish between positive and negative samples \cite{Endo2021,WangJ2022, Nazarov2022,Jeong2023}. 
For example, Wang et al. \cite{WangJ2022} introduced a multi-label contrastive loss to support their cross-modal prototype model in learning class-related and informative cross-modal patterns. CheXbert was then implemented to generate a pseudo label for each image-text pair, considering a sample as positive if it shares at least one label and negative if it does not share any common labels. 
Jeong et al. \cite{Jeong2023} generated hard negative samples to encourage their model to learn more robust representations. Instead of K-means sampling, they applied the CheXbert \cite{Smit2020} labeller to generate embeddings and create negative and ``hard'' negative samples by examining their Manhattan distance, with ``hard'' samples having smaller distances. Their model is trained against the hardest negative sample. 

As a rule-based labeller, an issue with CheXpert is that it is non-differentiable which prevents it being used with techniques such as backpropagation \cite{backpropagation}. To be able to directly train their model to produce clinically accurate reports, Lovelace and Mortazavi \cite{Lovelace2020} developed a method to differentiably extract clinical observations from generated reports and utilised this differentiability to fine-tune their model in a supervised fashion. They trained CNN and long short-term memory (LSTM) models to approximate the CheXpert labeler \cite{CheXpert} during training.

\subsection{Triplet Loss}
Although not as commonly applied as contrastive loss, triplet loss \cite{tripletloss} has been used in ARRG for training the model encoder, and as an auxiliary task to improve a model's performance. While both triplet loss and contrastive loss compare samples, the mechanism is a little more nuanced. Triplet loss, in particular, incorporates an additional sample to form a triplet, which includes a positive and negative sample for contrasting against the anchor. The aim is then to minimize the distance between the anchor and positive samples while maximizing the distance between the anchor and the negative samples to create more distinct clusters within the latent space.

Wang et al. \cite{WangZ2021} proposed a framework that uses triplet loss to help boost its performance during training. The framework consists of two branches where the main branch tackled the standard report generation task and the auxiliary branch performed image-text matching with triplet loss to learn text correlated visual features. Their approach trained the branches in an alternating manner 
and sought to boost training in three ways: (i) the feature extractor was used by both branches, providing twice as many training steps; (ii) the report encoder of the auxiliary branch is also used by the main branch to evaluate and minimise the feature-level loss between the generated and the ground-truth reports;
(iii) the reports generated by the main branch were used as ``hard'' negative samples while training the auxiliary branch, which as the network trained, became semantically closer to the reference report.

Two types of triplets were applied in Yang et al. \cite{Yang2022b}'s contrastive triplet network (CTN). The first triplet was the traditional positive, anchor, negative sample approach, while the second type of triplets reversed this, with abnormal cases as the anchor and positive class, and normal cases as the negative class. These triplets were used during training, where the visual embedding from the encoder and the semantic embedding of the reports were contrasted to encourage the similarity between positive and anchor cases and the dissimilarity between negative and anchor cases.


\subsection{Reinforcement Learning}\label{sec:reinforcement} Methodologies utilising reinforcement learning (RL) have been effective within the ARRG domain \cite{Delbrouck2022, Qin2022, WuY2023} with most contributions defining rewards for agents to optimise.

Two reinforcement rewards were proposed by Miura et al. \cite{Miura2021}: $fact_{ENT}$, or exact entity match reward, sought to capture factual completeness by extracting entities from both the generated and corresponding ground truth report, after which the reward function checks for matching entities, and $fact_{ENTNLI}$ or entailing entity match reward extended $fact_{ENT}$ by giving weight to the context of entities within sentences rather than just checking for an occurrence. The reward searched for contradictions between the two reports, such as a negation changing the meaning of the sentence. With $fact_{ENTNLI}$ a match only occurs if both sentences have the same semantic meaning, providing a more robust metric than $fact_{ENT}$.

``RadGraph reward'' was introduced by Delbrouck et al. \cite{Delbrouck2022} which involved the use of three RL rewards. The approach harnessed the PubMedBERT model \cite{PubMedBERT}, trained on the RadGraph dataset \cite{Radgraph}, to extract entities and relations from reports, where entities represent anatomical and clinical observations and relations denote the connections between the entities, such as ``suggestive of'' or ``located at''. For each case the PubMedBERT model created two semantic graphs, one generated from their ARRGs model's output and the other from the associated ground truth report. Their rewards evaluated the similarities between the two generated graphs, comparing the entities and relations that were extracted. By using RL, a simpler model was able to be quickly trained and provided equivalent performance to more complex architectures.

Imbalanced token distributions naturally exist within radiology reports, and this disparity has been found to cause ARRG models to overfit on frequent tokens, leading to lower performance on uncommon tokens. To combat this, Wu et al. \cite{WuY2023} reduced token frequency effects by adopting RL to dynamically adapt the token set during training. 
Their reward consisted of two components, a text generation reward and an imbalanced evaluation reward which broke down the comparison of performance differences into four groups based on how often the tokens appear. The F1 score was then calculated for each group of tokens, balancing performance to ensure the tokens from the groups are given more equal weights. 

Other works of note that utilise RL have generally employed NLP metrics as their reward \cite{WangZ2021, Qin2022, WangZ2022, Nishino2022} to reward lexical similarity to a reference report. 


\section{Architecture}
\label{subsec:transformers}
The architecture of ARRG models determine how they process and interpret radiographic images and reports 
and generate radiology reports from unseen radiographic images. To date, there are two dominant architectures used, encoder-decoder models and retrieval models. This section discusses various avenues of research which has sought to provide novel contributions to such architectures. Figure~\ref{fig:architectures} illustrates the three main architectures of encoder-decoder networks that are utilised within ARRG.

\subsection{CNN-LSTM}
Table~\ref{tbl:architectures} demonstrates that encoder-decoder models are the most frequently adopted architecture type for the ARRG task. Within this architecture, the encoder extracts features from the input radiographic image which are in turn used by the decoder to generate a free text radiology report. As discussed in Section~\ref{sec:background}, early research employed CNN-RNN or CNN-LSTM models, which are still in use in more current works \cite{ZhangY2020, WangZ2021, Liu2021b, WangS2022, Nishino2022, Gajbhiye2022, Wu2022, WangF2022, Kaur2023, Akbar2023}, which can be seen in Figure~\ref{fig:architectures}(A). The novelty of these approaches generally comes from 
additionally proposed modules or strategies within their networks.

The following approaches harnessed pretrained CNNs for their encoder and a Hierarchical LSTM for their decoder composed of two layers: a sentence LSTM and a word LSTM. The sentence LSTM generates topics for each sentence which are transferred to a word-LSTM to generate sentences on a word by word basis, using the topics as inspiration. Both Zhang et al. \cite{ZhangY2020} and Wang et al. \cite{WangS2022} employed pretrained DenseNet-121 \cite{Huang2017} CNN encoders 
with their novelty being grounded in their graph convolutional neural network modules. Wang et al. \cite{WangS2022} also introduced a knowledge graph within their architecture which assists the generation process by providing medical knowledge to the model.

A VGG19 encoder was employed by Kaur \& Mittal \cite{Kaur2023} for extracting image features and their distinctive contribution lies in the reduction of model size and complexity achieved through parameter pruning. Wang et al. \cite{WangZ2021} harnessed a ResNet-101 \cite{ResNet} for their encoder48 with their novelty being introduced through their self-boosting training regime. Liu et al. \cite{Liu2021b} employed ResNet-50 as their encoder and introduced a contrastive attention module as part of their contribution.

Other works implemented a Bi-Directional LSTM (Bi-LSTM) for their decoder where the two layers process information in opposite directions. To improve the performance of their model, Wang et al. \cite{WangF2022} proposed a hybrid model which unified retrieval and sentence generation approaches. Their adaptive generator selected between retrieving a sentence from a corpus of reports or utilising their LSTM decoder to generate sentences. Gajbhiye et al. \cite{Gajbhiye2022} integrated an adaptive multilevel multi-attention module to learn the association between visual and textual features. Their Bi-LSTM decoder explored these embeddings to generate a radiological report.

\begin{table*}
\scriptsize
    \centering
    \begin{tabular}{llllll} \hline 
         \textbf{~~~~~~~~~~~~Model}  &\textbf{Year}&  \textbf{Feature Extractor}&  \textbf{Model Type}& \textbf{Model Architecture}&\textbf{Model Output}\\ \hline 
          R2Gen \cite{Chen2020} &2020&ResNet& Encoder-Decoder& Transformer&Findings
\\ \hline            
          Syeda-Mahmood et al. \cite{SyedaMahmood2020} &2020&CNN&Retrieval&CNN&Both
\\ \hline 
          Zhang et al. \cite{ZhangY2020} &2020&DenseNet&Encoder-Decoder& GraphCNN-LSTM&Both
\\ \hline 
          Lovelace \& Mortazavi \cite{Lovelace2020} &2020&Densenet&Encoder-Decoder& Transformer&Findings
\\ \hline 
          Boag et al. \cite{Boag2020} &2020&Densenet&Various& Various&Findings
\\ \hline 
          \rowcolor[gray]{0.95}
          VTI \cite{Najdenkoska2021}  &2022&Densenet& Encoder-Decoder&  CNN-LSTM&Both
\\\hline
          \rowcolor[gray]{0.95}
          CMCL \cite{Liu2021}  &2021&ResNet& Encoder-Decoder&  Various&Findings
\\\hline
          \rowcolor[gray]{0.95}
          CA \cite{Liu2021b}  &2021&ResNet& Encoder-Decoder&  Various&Findings
\\\hline
          \rowcolor[gray]{0.95}
          Nooralahzadeh et al. \cite{Nooralahzadeh2021}  &2021&Densenet& Encoder-Decoder&  Transformer&Findings
\\\hline
          \rowcolor[gray]{0.95}
          Yan et al. \cite{Yan2021}  &2021&ResNet& Encoder-Decoder&  Transformer&Findings
\\\hline
          \rowcolor[gray]{0.95}
          AlignTransformer \cite{You2021} &2021&ResNet&Encoder-Decoder&   Transformer&Findings
\\\hline
          \rowcolor[gray]{0.95}
          PPKED \cite{LiuF2021} &2021& ResNet& Encoder-Decoder& Transformer&Findings
\\\hline
          \rowcolor[gray]{0.95}
          KGAE \cite{liu2021autoencoding} &2021& ResNet& Encoder-Decoder& Transformer&Findings
\\\hline
          \rowcolor[gray]{0.95}
          R2GenCMN \cite{Chen2021} &2021& ResNet& Encoder-Decoder&  Transformer&Findings
\\\hline
          \rowcolor[gray]{0.95}
          Nguyen et al. \cite{Nguyen2021} &2021& ResNet& Encoder-Decoder&  Transformer& Findings
\\\hline
          \rowcolor[gray]{0.95}
          MedWriter \cite{YangX2021} &2021& Densenet& Various&  CNN-LSTM&Findings
\\\hline
          \rowcolor[gray]{0.95}
          CXR-RePaiR \cite{Endo2021}  &2021& ResNet / ViT & Retrieval&  ResNet / ViT&Both
\\\hline
          \rowcolor[gray]{0.95}
          CDGPT2 \cite{Alfarghaly2021}  &2021& ResNet& Encoder-Decoder&  CNN-Transformer&Both
\\\hline
          \rowcolor[gray]{0.95}
          Wang et al. \cite{WangZ2021}  &2021&ResNet&Encoder-Decoder&  CNN-LSTM&Findings
\\ \hline 
          \rowcolor[gray]{0.95}
          Muira et al. \cite{Miura2021}  &2021&Densenet& Encoder-Decoder&  Transformer&Findings
\\ \hline
          \rowcolor[gray]{0.95}
          Medical-VLBERT \cite{LiuG2021} & 2021 &DenseNet&Encoder&BERT&Findings \\ \hline
          \rowcolor[gray]{0.9}
          Zhao et al. \cite{Zhao2022} &2022& ResNet& Encoder-Decoder&  Transformer&Findings
\\ \hline 
          \rowcolor[gray]{0.9}
          Wang et al. \cite{Wang2022}  &2022&ViT&Encoder-Decoder&   Transformer&Findings
\\\hline    
          \rowcolor[gray]{0.9}
          Radgraph Reward  \cite{Delbrouck2022}  &2022&Densenet& Encoder&  BERT&Findings
\\\hline       
          \rowcolor[gray]{0.9}
          Yang et al. \cite{Yang2022b}  &2022&ResNet& Encoder-Decoder&  Transformer&Findings
\\\hline
          \rowcolor[gray]{0.9}
          Qin and Song \cite{Qin2022}  &2022&ResNet& Encoder-Decoder&  Transformer&Findings
\\\hline
          \rowcolor[gray]{0.9}
          Yang et al. \cite{Yang2022}  &2022&ResNet& Various&  CNN-Transformer&Findings
\\\hline
          \rowcolor[gray]{0.9}
          VTI-TRS \cite{Najdenkoska2022}  &2022&Densenet& Encoder-Decoder&  CNN-LSTM/Transformer&Both
\\\hline
          \rowcolor[gray]{0.9}
          XPRONET \cite{WangJ2022}  &2022&ResNet& Encoder-Decoder&  CNN-Transformer&Findings
\\\hline
          \rowcolor[gray]{0.9}
          CMCA \cite{Song2022}  &2022&Densenet& Various&  CNN-Transformer&Findings
\\\hline
          \rowcolor[gray]{0.9}
          MaKG \cite{YanS2022}  &2022&Densenet& Encoder-Decoder&  Transformer&Findings
\\\hline
          \rowcolor[gray]{0.9}
          Zhang et al. \cite{Zhang2022} &2022&Densenet& Encoder-Decoder&  CNN-Transformer&Findings
\\\hline
          \rowcolor[gray]{0.9}
          Nazarov et al. \cite{Nazarov2022}  &2022&Various& Encoder-Decoder&  Various&
Impression
\\\hline
          \rowcolor[gray]{0.9}
          MATNet \cite{Shang2022} &2022&Densenet& Encoder-Decoder&  Transformer&
Findings
\\\hline      
          \rowcolor[gray]{0.9}
          Dalla Serra et al. \cite{DallaSerra2022} &2022&ResNet& Encoder-Decoder&  Transformer&
Findings
\\\hline       
          \rowcolor[gray]{0.9}
          SGF \cite{Li2022}  &2022&ResNet& Encoder-Decoder&  Transformer&
Findings
\\\hline       
          \rowcolor[gray]{0.9}
          JPG \cite{You2022}  &2022&ResNet& Encoder-Decoder&  Transformer&
Findings
\\\hline
          \rowcolor[gray]{0.9}
          MedViLL \cite{Moon2022} &2022&ResNet&Encoder&BERT&Both \\\hline
          \rowcolor[gray]{0.9}
          AMLMA \cite{Gajbhiye2022}  &2022&Densenet& Encoder-Decoder& CNN-LSTM&Findings
\\\hline
          \rowcolor[gray]{0.9}
          Relation-paraNet \cite{WangF2022} &2022&VGG19 / Densenet& Encoder-Decoder&  CNN-LSTM&
Findings
\\\hline
          \rowcolor[gray]{0.9}
          Clinical-BERT \cite{YanB2022} &2022&Densenet& Encoder& BERT&
Both
\\\hline
          \rowcolor[gray]{0.9}
          RepsNet \cite{Tanwani2022} &2022&ResNeXt& Encoder&  BERT&
Findings
\\\hline
          \rowcolor[gray]{0.9}
          MSAT \cite{WangZ2022} &2022&CLIP& Encoder-Decoder&  Transformer&
Findings
\\\hline
          \rowcolor[gray]{0.9}
          Wang et al. \cite{WangS2022}  &2022&Densenet& Encoder-Decoder&  CNN-LSTM&Findings
\\\hline
          \rowcolor[gray]{0.9}
          CoPlan \cite{Nishino2022} &2022&CNN& Encoder-Decoder&  CNN-LSTM&Findings
\\\hline
          \rowcolor[gray]{0.9}
          Leonardi et al. \cite{Leonardi2022} &2022&CNN&Encoder-Decoder&CNN-Transformer&Findings \\ \hline
          \rowcolor[gray]{0.9}
          Yan et al. \cite{Yan2022} & 2022&ResNet&Encoder-Decoder&Transformer&Findings \\ \hline
          \rowcolor[gray]{0.9}
          Wang et al. \cite{Wang2022b} & 2022 &ResNet&Encoder-Decoder&Transformer&Findings \\ \hline
          \rowcolor[gray]{0.9}
          MRCL \cite{Wu2023} & 2022& ResNet&Encoder-Decoder&CNN-LSTM/BERT &Both \\ \hline
          \rowcolor[gray]{0.9}
          Boecking et al. \cite{Boecking2022} & 2022&ResNet&Encoder&BERT&Findings \\ \hline
          \rowcolor[gray]{0.9}
          \ps{DeltaNet} \cite{Wu2022} & \ps{2022}&\ps{ResNet}&\ps{Various}&\ps{CNN}&\ps{LSTM} \\ \hline
          \rowcolor[gray]{0.9}
          TrMRG \cite{Mohsan2023}  &2023&CNN&Encoder-Decoder&   Transformer&
Both
\\ \hline
          \rowcolor[gray]{0.85}
          CheXPrune \cite{Kaur2023}  &2023&VGG19& Encoder-Decoder&  CNN-LSTM&
Both
\\\hline
          \rowcolor[gray]{0.85}
          Kim et al. \cite{Kim2023}  &2023&Densenet& Encoder-Decoder&  Transformer&
Findings
\\\hline
          \rowcolor[gray]{0.85}
          MKCL \cite{Hou2023b}  &2023& ResNet&Encoder-Decoder& Transformer&
Findings
\\\hline
          \rowcolor[gray]{0.85}
          CvT-212DistilGPT2 \cite{Nicolson2023}  &2023& Various&Encoder-Decoder&  Various&
Findings
\\\hline
          \rowcolor[gray]{0.85}
          ASGK \cite{Li2023} &2023&Densenet& Encoder-Decoder& CNN-Transformer&
Findings
\\\hline
          \rowcolor[gray]{0.85}
          BioViL-T \cite{Bannur2023} &2023&ResNet&Encoder-Decoder&   Transformer&
Findings
\\ \hline
          \rowcolor[gray]{0.85}
          ORGAN \cite{Hou2023}  &2023&ResNet&Encoder-Decoder&   Transformer&
Findings
\\ \hline 
          \rowcolor[gray]{0.85}
          X-REM \cite{Jeong2023}  &2023&ALBEF \cite{ALBEF}& Retrieval&  ALBEF \cite{ALBEF}&
Both
\\\hline
          \rowcolor[gray]{0.85}
          MRCL \cite{Wu2023}  &2023&CNN& Encoder-Decoder& CNN-LSTM/Transformer&
Both
\\\hline
          \rowcolor[gray]{0.85}
          KiUT \cite{Huang2023}  &2023&ResNet& Encoder-Decoder&  Transformer&
Findings
\\\hline
          \rowcolor[gray]{0.85}
          TIMER \cite{WuY2023}  &2023& ResNet& Encoder-Decoder&  Transformer&
Findings
\\\hline
          \rowcolor[gray]{0.85}
          Yang et al. \cite{Yang2023} &2023& ResNet& Encoder-Decoder&  Transformer&
Findings
\\\hline
          \rowcolor[gray]{0.85}
          Kale et al. \cite{Kale2023} &2023& ResNet& Encoder-Decoder&  Transformer&
Findings
\\\hline
          \rowcolor[gray]{0.85}
         METransformer \cite{Wang2023} &2023 & ViT & Encoder-Decoder& Transformer & Findings\\\hline
           \rowcolor[gray]{0.85}
          ATAG \cite{YanS2023} &2022&Densenet& Encoder-Decoder&  CNN-LSTM/Transformer&
Both
\\\hline
          \rowcolor[gray]{0.85}
          Selivanov et al. \cite{Selivanov2023} & 2023 & CNN & Encoder-Decoder& CNN-LSTM/Transformer & Findings\\\hline
          \rowcolor[gray]{0.85}
          RGRG \cite{tanida2023} &2023 & Faster R-CNN& Encoder-Decoder & Faster RCNN-Transformer & Findings\\\hline
          \rowcolor[gray]{0.85}
          Gu et al. \cite{Gu2023} &2023&CNN&Encoder-Decoder&CNN-LSTM& Findings\\\hline
          \rowcolor[gray]{0.85}
          \ps{UAR} \cite{Li2023b} &\ps{2023}&\ps{dVAE}&\ps{Encoder-Decoder}&\ps{dVAE-Transformer}& \ps{Findings}\\\hline
          \rowcolor[gray]{0.85}
          \ps{Dalla Serra et al.} \cite{DallaSerra2023} &\ps{2023}&\ps{Faster R-CNN} &\ps{Encoder-Decoder}&\ps{Faster RCNN-Transformer}& \ps{Findings}\\\hline
          \rowcolor[gray]{0.85}
          \ps{Zhu et al.} \cite{Zhu2023} &\ps{2023}&\ps{CNN} &\ps{Encoder-Decoder}&\ps{CNN-Transformer}& \ps{Findings}\\\hline
          \rowcolor[gray]{0.85}
          \ps{Nguyen et al. \cite{Nguyen2023}} &\ps{2023}&\ps{ResNet} &\ps{Various}&\ps{CNN-Transformer}& \ps{Findings}\\\hline
    \end{tabular}
    \caption{{Architectures employed by ARRG Research with a breakdown of feature extractors, conceptual model type and architectures of the proposed models. Models are designed to generate content for either the findings section, the impression section, or both sections of a radiology report.}}
    \label{tbl:architectures}
\end{table*} 

Variational Topic Inference (VTI) was proposed by Najdenkoska et al. \cite{Najdenkoska2021} who introduced latent variables in the form of a set of topics to guide the sentence generation process. Their model used a Hierarchical LSTM for sentence generation. Upon scrutinising their findings, they observed instances where sentences exhibited missing words which they attributed to the LSTM's handling of long-term dependencies. Building on this work, Najdenkoska et al. \cite{Najdenkoska2022} proposed VTI-TRS, taking their original proposal and adopted the transformer architecture. Each attention head created a specific topic representation, which is used to create a sentence within the generated report. Within their new research they trained models employing either an LSTM or a transformer decoder. Ablation studies revealed that the transformer decoder yielded the most optimal results. 

\subsection{Encoder-decoder Transformers}
Many recent works have leveraged the conventional transformer architecture \cite{Vaswani}, which was initially evaluated for NLP tasks such as translating text from one language to another (see Table~\ref{tbl:architectures}). These approaches have often adopted a CNN feature extractor to manipulate the radiographic image into an appropriate input for the traditional transformer encoder. An illustration of a typical architecture of this type can be seen in Figure~\ref{fig:architectures}(B). As depicted in Figure~\ref{fig:architectures}(C), other researchers \cite{Wang2022, Shang2022, Mohsan2023}  utilised the vision transformer \cite{Dosovitskiy2021} to integrate the image directly into their model.

A radiology report goes beyond identifying anatomical structures and pathological findings, and also provides detailed information about the size, location, and severity of any findings. Syeda-Mahmood et al. \cite{SyedaMahmood2020} sought to extract this additional information from the radiographic images, denoting these findings as fine finding labels (FFL). 
They trained two networks, one to extract traditional labels and another to extract FFLs, the features from which were then consolidated into a pattern vector. The closest matching report was then selected by finding the most semantically similar report vector within their database.

Rather than simply extracting visual features from the images, some works have introduced an intermediate step to the report generation process. Nooralahzadeh et al. \cite{Nooralahzadeh2021}'s first step employed a Meshed-Memory Transformer \cite{meshtransformer} for generating high level textual concepts to be reformed into finer and more fluent reports by their second step, a transformer language model called BART \cite{BART} which contains a BERT \cite{BERT} encoder and GPT-2 \cite{Radford2019} decoder. Dalla Serra et al. \cite{DallaSerra2022} also divided the ARRG task into two components: factual triple extraction and report generation. To train their model, ground truth triples in the form $(entity1, relation, entity2)$ were generated by merging outputs from RadGraph \cite{Radgraph} and ScispaCy \cite{ScispaCy}, which were exploited, along with the associated clinical indications, to train their transformer encoder for triplets production. Their report generation stage of the network incorporated the triplets, in conjunction with the input radiographic image, to generate the radiology report.

{As an alternative to extracting image-level features, Tanida et al. \cite{tanida2023}'s approach, Region-Guided Radiology Report Generation, focused on leveraging a FasterRCNN to extract 29 distinct anatomical regions of the chest. Their region selection module then predicted (by binary classification) whether a sentence should be generated for each anatomical region. Their decoder, a transformer-based language model that was pretrained on medical abstracts, then generated sentences for each of the salient anatomical regions. }
Similarily, Wang et al. \cite{WangY2022} proposed AGFNet, which employed a global and region feature extractor and a self-adaptive fusion gate to improve their ARRG model. Their feature extractor produced region features by utilising a Faster R-CNN head and region of interest pooling. These region features, along with the global features, were processed by their self-adaptive fusion gate which used multi-head attention as its main mechanism. \ps{An image segmentation approach was adopted by Nimalsiri et al. \cite{Nimalsiri2023} within their transformer-based architecture, incorporating a U-Net architecture \cite{Ronneberger2015} to first extract lung images, from which image features were then derived. Wang et al \cite{WangH2023} clustered discrete grid features with similar semantics into groups to obtain key region features with the vector of locally aggregated descriptors method \cite{Arandjelovic2018} being used to determine similar regions.}

The Knowledge-injected U-Transformer is a framework developed by Huang et al. \cite{Huang2023} which focused on extracting and distilling multi-level information and multiple injected knowledge. Their model contained three main components: a cross-modal U-Transformer, the injected knowledge distiller, and the region relationship encoder. Their cross-modal U-Transformer is the main architecture of the model which contains connections between each level of the transformer, similar to skip connections in U-Nets \cite{Ronneberger2015}.

Wang et al. \cite{Wang2023} incorporated ``expert" tokens (a learnable embedding) as queries in their expert bilinear attention module, and employed these tokens in both their transformer encoder and decoder. 
For their encoder, each expert token interacted with both vision tokens and other expert tokens to learn to attend different image regions for image representation to capture complementary information by an orthogonal loss that minimizes their overlap. In their decoder, each attended expert token guided the cross-attention between input words and visual tokens, influencing the generated report. A metrics-based expert voting strategy generated the final report.

Exploiting the fact that the majority of sentences within a radiology report describe normal findings, Kale et al. \cite{Kale2023} developed a novel method that incorporated a ``normal'' template report. Their ResNet-50 visual extractor was trained to predict MeSH or CheXpert labels from IU-Xray and MIMIC-CXR, respectively, aiming to identify abnormal areas in the input image. Should any of these regions be deemed abnormal, their transformer-based text generator generated a sentence describing the observed pathology. Subsequently, their replacement module, a BERT \cite{BERT} architecture, determined whether to replace a ``normal'' sentence or append the sentence as an additional statement within the template report.

\begin{figure}[ht]
\centerline{\includegraphics[scale=.425]{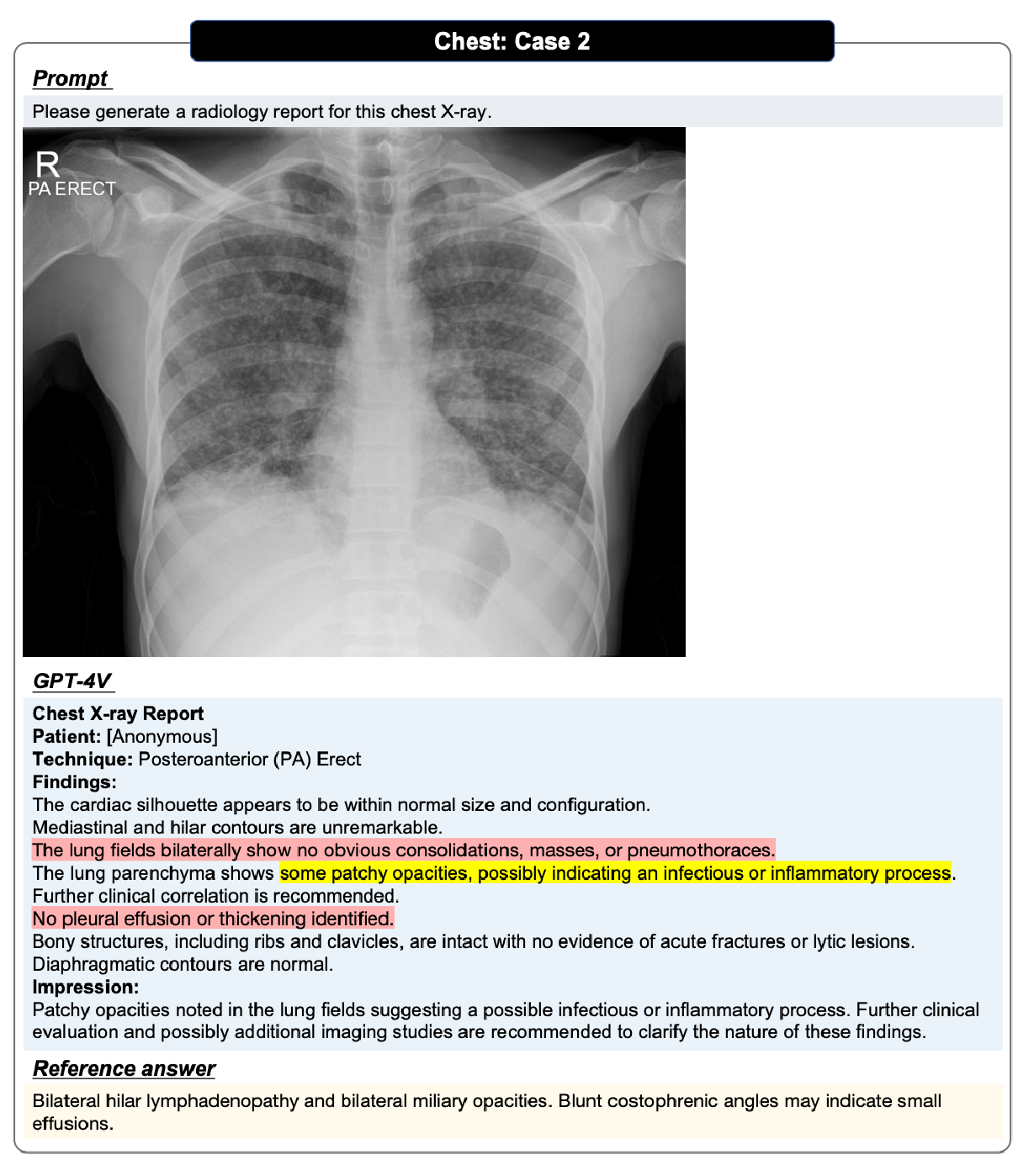}}
\caption{A sample prompt and GPT-4V's responses from Chaoyi et al.'s \cite{Chaoyi2023} qualitative analysis. The chest x-ray demonstrates a case of military tuberculosis. GPT-4V was requested to generate a radiology report. The resulting findings section shows that while GPT-4V is capable of conducting a comprehensive analysis of the chest x-ray, its descriptions and assessment of the pathology are not accurate.}
\label{fig:GPT4V}
\end{figure}

\subsection{Pretrained Language Models}\label{subsec:llms} Due to their impressive performance in a variety of NLP tasks \cite{llama, falcon}, there has been a surge of interest investigating pretrained language models in the ARRG domain, beginning with models such as BERT \cite{BERT} and GPT \cite{Radford2019} which are based on the transformer encoder and decoder respectively. 

{BERT-related architectures are currently the most frequently used pretrained language model within the ARRG domain, either as a stand alone model \cite{Delbrouck2022, Moon2022, YanB2022} or as a module within their architecture \cite{Nooralahzadeh2021, WangZ2021, Tanwani2022, WuX2023, Jeong2023, Kale2023}.} 

{Approaches that use BERT as a stand alone model seek to generate the report in an autoregressive manner \cite{Delbrouck2022, Moon2022, YanB2022}. Being an encoder-based model, it inherently lacks generative capabilities, but it can execute this task by treating it as a sequential classification task. Generating reports in this fashion is initiated with a ``MASK'' token, a placeholder token where the model is predicting the most likely word. Once predicted, another ``MASK'' token is added to the end of the sequence until the model forecasts a ``SEP'' token, indicating the conclusion of the sequence. Moon et al. \cite{Moon2022} trained their model using masked language modelling (MLM), where part of a report was masked with a ``MASK'' token with the model being tasked to predict the token, and image-report matching which was used to encourage the model to learn both visual and textual features to predict whether an image and a report are a matching pair.}

Approaches that integrated BERT into their network applied the module in a variety of ways. Wang et al. \cite{WangZ2021} implemented Sentence-BERT \cite{sentence-bert} in their self-boosting network, with the module performing an auxiliary task of image-text matching. Kale et al. \cite{Kale2023} trained a BERT-based multilabel text classifier to identify the relevant sentences within their ``normal'' template report so they can be replaced with the generated sentences that describe pathologies that are present within the input radiographic image. Jeong et al. \cite{Jeong2023} adopted the Align Before Fuse (ALBEF) model \cite{ALBEF} for their retrieval based approach to the ARRG task. ALBEF is an encoder model with three components: a vision transformer \cite{Dosovitskiy2021} for encoding the radiographic images and two BERT models are used for their text and multimodal encoders. 

The introduction of large language models (LLMs), particularly GPT-3 \cite{Brown2020}, marked quite a significant shift due to the models impressive generalisation in zero-shot settings, which is due to an increase in the number of parameters and training data. While ARRG research with generative LLMs is not yet as widespread as research focusing on encoder-related LLMs, the availability of publicly available foundational models such as LLaMA \cite{llama} and Falcon \cite{falcon} have enhanced their accessibility to the research audience. As such, it is anticipated that these foundational models will be investigated for the ARRG task in the near future.

Selivanov et al. \cite{Selivanov2023} sought to improve the quality of the encoder-decoder generated radiology reports by using the GPT-3 language model. Their model consists of two parts, show, attend and tell (SAT) \cite{SAT} and GPT-3 \cite{Brown2020}. They proposed two approaches for combining the models. The first model forced the models to learn a joint word distribution by concatenating their outputs and using a feed-forward network to obtain a final score for each token. Their second approach utilised them in a stacked approach where SAT took the input radiographic image and created an initial report after which GPT-3 refined the initial report to create a more fluent report, with this stacked approach proving more effective.

Chaoyi et al. \cite{Chaoyi2023} performed qualitative analysis of OpenAI's latest model, GPT4-V \cite{ChatGPT}, which supports multimodal inputs such as images. An example case study can be seen in Figure~\ref{fig:GPT4V}. The study evaluated the models capacity for medical diagnosis in various radiology modalities. They assessed GPT-4V's efficacy at identifying anatomical structures, disease diagnosis and the generation of reports for various parts of the body, including the chest. Zero shot learning was used with images obtained from Radiopaedia \footnote{https://radiopaedia.org/}.
GPT-4V was observed to be proficient at identifying between image modalities and anatomy, but faces challenges in disease diagnosis and report generation. Due to the limitation that OpenAI have not yet released an API for this model, a quantitative use is not yet possible. 

Other papers of significance where various approaches to ARRG were introduced by way of pretrained language models are \cite{Alfarghaly2021, LiuG2021, YanB2022, Delbrouck2022, Boecking2022, Bannur2023, Nguyen2023}.


\subsection{Attention Modules}
Due to the fine-grained differences within radiographic images, researchers have often sought to develop novel attention mechanisms. These attention modules aim to more effectively capture the subtle changes within the images, ultimately enhancing the robustness of features and improving the decoder's ability to generate accurate reports. Various different solutions for the attention mechanism have been developed, for example \cite{Gajbhiye2022, Song2022, WangZ2022, Yan2022}.

{To effectively obtain the visual features of pathologies, Song et al. \cite{Song2022} proposed a cross-modal contrastive attention (CMCA) model which compared the input features extracted by their encoder against the most similar historical case retrieved from their database, where each case contained global and spatial visual features along with the corresponding report. 
The contrastive features created by their CMCA module were then used, along with the retrieved report, to guide the report generation process.} 

Adaptive Multilevel Multi-Attention was developed by Gajbhiye et al. \cite{Gajbhiye2022} which contained two separate attention mechanisms, adaptive visual semantic attention (AVSA) and visual based language attention (VBLA). AVSA adaptively selected the weight of visual global information and semantic patterns to attend to their local visual features while VBLA module was responsible for learning the correlation and inter-dependency between the consecutive words and sentences. Align Hierarchical Attention was proposed by You et al. \cite{You2021} which was integrated into their AlignTransformer model to help correspond visual features with disease tags.

{Gu et al. \cite{Gu2023} leveraged the multi-perspective nature of radiology images to introduce cross-view attention. This  operated through two branches, a single-view branch preserving single-view features, and a multi-view branch combining features extracted from both frontal and lateral images using squeeze-excitation attention to create multi-view features. The single-view features, multi-view features and medical concepts (MTI tags from IU-Xray dataset) were then used by their Medical Visual-Semantic LSTM to help generate a radiology report.}

Memory-augmented Sparse Attention (MSA) was proposed by Wang et al. \cite{WangZ2022} with the aim of capturing second or even higher-order interactions by incorporating bilinear pooling into self-attention. In their MSA module, global features were considered as the query, while local features served as the key and value. Additional "memory slots" were  also introduced to the keys and values, facilitating the encoding and collection of features from all preceding processes with the intention of  generating long reports more accurately. {Sparse attention was also utilised by Yan et al. \cite{Yan2022} who developed a two-tiered approach consisting of region sparse attention and word sparse attention. To enhance semantic alignment in their model, they harnessed the MeSH labels from the MIMIC-CXR dataset. Their region sparse attention acquired region features corresponding to each MeSH label, while their word sparse attention assigned increased weights to the MeSH words present in a report.}

\begin{figure*}[!ht]
\centerline{\includegraphics[scale=0.325]{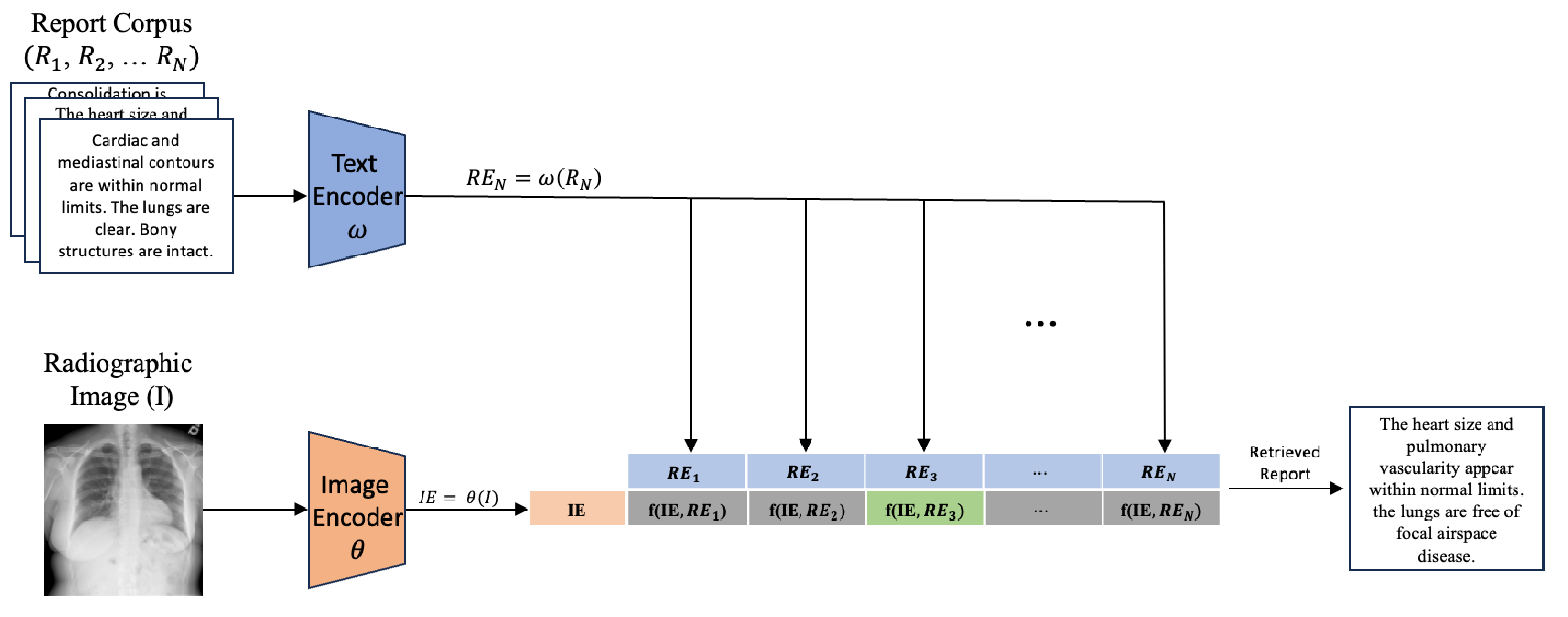}}
\caption{A representation of the retrieval based architecture for the ARRG task where, given an input radiographic image, the most similar report, which is calculated through a distance function, is retrieved from a collection of reports. Several researchers have a corpus that contains both sentences and reports, for example \cite{Endo2021, YangX2021, WangF2022, Jeong2023}.}
\label{fig:retrieval}
\end{figure*}

\subsection{Retrieval-based methods}
An alternative solution to the ARRG task is the retrieval approach where models consider a corpus of radiological knowledge to find similar reports or sentences to create their reports. An illustration of a retrieval-based method can be seen in Figure~\ref{fig:retrieval}. Yang et al. \cite{YangX2021} developed MedWriter, which incorporated a hierarchical retrieval mechanism comprising,  (i) a visual-language retrieval module operated on the report level, using visual features to source the most semantically similar report template in their retrieval pool, (ii) a language-language retrieval module working on the sentence level, where a set of candidate sentences are retrieved from the retrieval pool that have potential to be the next sentence, and (iii) a hierarchical-LSTM decoder fusing the features from the previous modules to generate their report. \ps{Wu et al. \cite{Wu2022} devised a comparable approach to Yang et al. \cite{YangX2021}, which either retrieved the image-report pair with the most similar visual features or, if prior imaging studies were available for a patient, then these were retrieved instead. Their model demonstrated the capability to process multiple prior imaging studies to assist with the generation of the current radiology report.}

{A retrieval-based approach called CXR-RePaiR was developed by Endo et al. \cite{Endo2021} which used a CLIP model with a retrieval corpus generated by collating the reports and sentences of the MIMIC-CXR dataset. The models were first trained on natural image-text pairs, followed by training on radiology image-report pairs where the model was trained to produce higher dot-product similarity values for a matching image-report pair. 
Their ablation studies found that sentence retrieval resulted in improved performance over report retrieval as sentence retrieval was considered to have a greater set of possible outputs to select from. \ps{Ramesh et al. \cite{Ramesh2022} retrained CXR-RePaiR \cite{Endo2021} on their version of MIMIC, CXR-PRO, finding their curated dataset improved their model's performance.} 
Similarly to Endo et al. \cite{Endo2021}, Jeong et al \cite{Jeong2023} utilised CLIP to improve the image-text matching score of their model, enabling a more accurate measure of the similarity of a chest X-ray image and radiology report for report retrieval. Their approach used a model called ALBEF \cite{ALBEF} which contained three fused encoders -- an image, text and a multimodal encoder. 
Instead of employing the dot-product, they applied cosine similarity as their distance measure function.}

{Wang et al. \cite{WangF2022} developed Relation-paraNet which unified retrieval and report generation through the use of their relational-topic encoder by taking image features and sentence embeddings of the previously generated sentences as input. The contextual topic embeddings, produced by their encoder, were used by their adaptive generator to determine whether to retrieve a template from their template database or generate a new sentence.}

Wu et al. \cite{WuX2023} proposed an LSTM module focused on creating the impression section of a report, and a BERT-based module that retrieved a set of appropriate sentences for the findings section. Both modules employed the same image encoder but each task was associated with its own final, fully connected layer to tailor the visual features specific to each task.


\section{Utilising Knowledge and Multiple Modalities}
\label{subsec:semantic}
Despite the improvements in models and architectures, e.g. the rewards harvested by the transformer, significant room for improvements remain in the ARRG domain.
Babar et al. \cite{Babar2021} investigated the efficacy of the encoder-decoder model and found that the evaluated ARRG models learnt an unconditioned model, not effectively attending to the image features during report generation. One of their recommendations was to utilise medical knowledge within the model.

Various ARRG researchers have developed methods to inject medical knowledge into their models \cite{Nishino2022, Gajbhiye2022, Li2022, Kim2023, Hou2023}. Kim et al. \cite{Kim2023} argued that due to the lack of prior knowledge, current approaches generated results containing references to non-existent prior examinations. They introduced comparison priors which were extracted from reports through a rule-based classifier. During training, reports were classified as being a first (negative) or a follow up examination (positive). They integrated their labeler into two publicly available models, R2Gen \cite{Chen2021} and M$^{2}$Tr \cite{meshtransformer}, improving both models scores on common NLG metrics by reducing the occurrence of incorrectly mentioned prior examinations. Wang et al. \cite{WangZ2022} created a Medical Concepts Generation Network. using the entities from RadGraph \cite{Radgraph} as medical semantic concepts, and selecting 768 concepts according to their occurrence frequency. Multi-label classification was then performed to determine which of the concepts were contained within the input radiographic image.


\subsection{Multimodal Inputs}\label{sssec:multimodality} 
\begin{figure}[t]
\centerline{\includegraphics[width=\columnwidth]{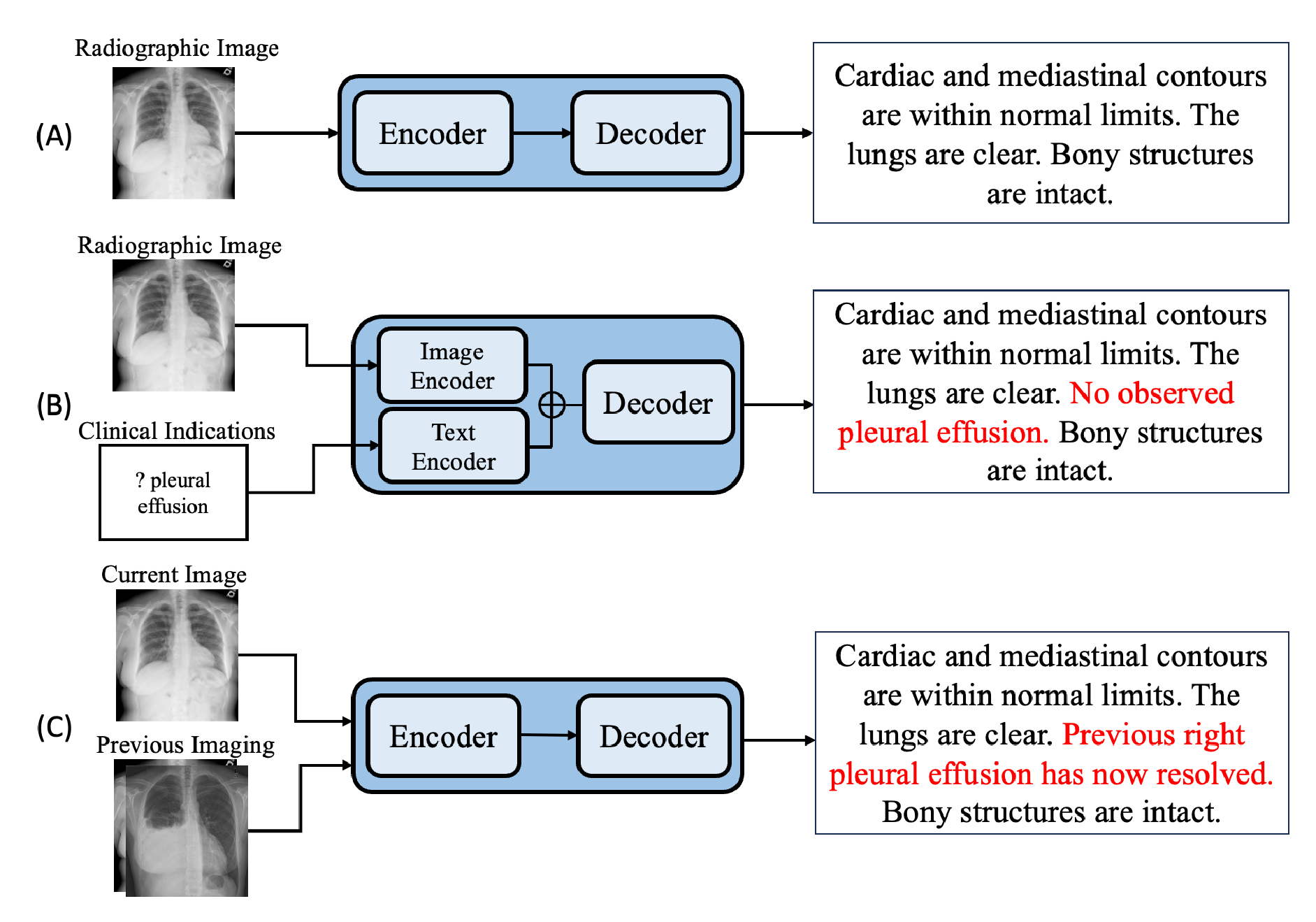}}
\caption{Illustration of  diverse forms of multimodal input used within ARRG research during inference. (A) demonstrates the most common, unimodal form of input to an ARRG model, the radiographic image. (B) represents models which adopted a multimodal input of images and text, with the text often being the clinical indications found within the image's corresponding reports \cite{Nguyen2021, Shang2022, DallaSerra2022}. (C) shows a model which implemented temporal information from prior radiographic images \cite{Bannur2023}.}
\label{fig:multimodal}
\end{figure}

ARRG is a multimodal task where both radiographic images and reports are used at various points throughout a contribution's pipeline. Reports have been used for implementing training approaches, such as CLIP \cite{Endo2021, Tanwani2022, Nazarov2022, WuX2023}, for example providing extra information to improve feature extraction \cite{SyedaMahmood2020, DallaSerra2022, DallaSerra2023, Mondal2023} or to create knowledge graphs \cite{liu2021autoencoding, Zhao2022, Hou2023}. In the inference phase however, the majority of ARRG architectures typically feature a unimodal encoder that takes radiographic images as input. Information such as clinical indications or prior examinations are often used by radiologists while reporting as they provide additional information about a patient's condition. To emulate this working style, some researchers have attempted to leverage this additional information to enhance their models during inference as this information is readily available at the time of reporting. Instances of research incorporating prior information during inference can be seen in Figure~\ref{fig:multimodal}.



Harnessing multimodal inputs, Nguyen et al. \cite{Nguyen2021} incorporated radiographic images and clinical details, which were fused into contextualised disease-related embeddings by their classification module. Similarily, Shang et al. \cite{Shang2022} also included  clinical referrals to provide their encoder with better learning capability. Dalla Serra et al. \cite{DallaSerra2022} used clinical indications to assist their feature extractor with creating more impactful features. Nguyen et al. \cite{Nguyen2023} leveraged clinical indications alongside the classifications from their ResNet encoder to assist their model with generating reports. \ps{Through the use of a transformer-based backbone, Mondal et al. \cite{Mondal2023} incorporated early fusion of radiology images and clinical indications to assist in classifying their samples.}

Rather than having separate visual and textul encoders, Jeong et al. \cite{Jeong2023} harnessed the ALBEF model \cite{ALBEF}, a multimodal encoder, to create multimodal representations of the radiographic image and reports. They utilised a trainable, shared matrix to learn new representations and used contrastive learning to help create more meaningful embeddings. Wang et al. \cite{Wang2022b} utilised their cross-modal fusion network to fuse visual features and disease tags of the current radiographic image.

Using the JLiverCT dataset, Nishino et al \cite{Nishino2022} introduced a temporal aspect to their model. This dataset consists of multi-phase CT examinations, where scanning was performed sequentially over a period of time, often tracking a contrast medium perfusing throughout a patients body. Their model exploited this temporal information to generate chronologically accurate reports. Instead of examining each sample individually, Bannur et al. \cite{Bannur2023} took advantage of the longitudinal nature of MIMIC-CXR by leveraging the images and reports from  patients' previous imaging studies as a secondary input to their model. These additional images and reports provided their model with prior information that enabled the model to generate sentences on the status of a noted pathology, and if its condition was improving or deteriorating. \ps{Dalla Serra et al. \cite{DallaSerra2023} also leveraged the temporal nature of MIMIC-CXR, using a Faster-R CNN object detector to extract anatomical tokens of the current and prior images. These tokens were aligned, concatenated and projected into joint representations before being passed alongside the clinical indications to their transformer based decoder to generate a report. Prior imaging was also used by Zhu et al. \cite{Zhu2023} to assist with generating the findings section of a radiology report with their transformer decoder being supplied with the longitudinal multimodal embeddings of the prior images and reports alongside the current image embedding.}

\subsection{Memory Modules} 
Some works have introduced memory mechanisms to provide a way of storing knowledge to boost a model's ability to generate clinically accurate reports \cite{Yang2023} 
Chen et al. \cite{Chen2020} employed a relational memory module in their transformer-based architecture to record information from previous generation processes and a novel memory-driven conditional layer normalization to integrate relational memory into their transformer architecture. This research contribution is one of the most frequently compared against models \cite{liu2021autoencoding, Wang2022, DallaSerra2022, WangZ2022, Li2022, Hou2023, Hou2023b, Wang2023, Huang2023, Kim2023, Li2023}. Chen et al. \cite{Chen2021} later expanded the R2Gen design by introducing the cross-modal memory network module. Rather than mapping between images and text directly from their encoded representations, they used a shared memory block as a medium to record the alignment between images and reports. 

Wang et al. \cite{WangJ2022} harnessed prior information from the training set of the MIMIC-CXR dataset 
to guide the initialisation of their semantic cross-modal prototype matrix. Their model, XPRONET, proposed a cross-modal prototype network, distilling image and textual information into single-model features. Central to their module was the prototype memory matrix, which contained image pseudo-labels. 


Yang et al. \cite{Yang2023} developed a memory module to be used as a knowledge base which was used to model the relationships between the image and textual features. Multi-head attention was the main mechanism used to update their knowledge bases' memory values. 

\subsection{Knowledge-based Graphs}\label{subsubsec:knowledgegraph} Knowledge graphs were popularised by Google in 2012 as a method to achieve a more intelligent search engine. Since its conception, it has gained popularity in deep learning research as a useful method to inject semantic knowledge into a model. A knowledge graph is essentially a structured base of knowledge with a directed graph structure where the nodes of the graph represent concepts while the edges of the graph represent semantic relations between the nodes. Due to this, a knowledge graph can well represent the relationship between nodes \cite{Zhang2022}, which has been utilised by many researchers to improve the performance of their models.

Amongst approaches which have manually built their knowledge graphs to hand pick the most pertinent categories, Zhang et al. \cite{ZhangY2020} constructed a chest abnormality graph based on prior knowledge. 
They selected 20 categories which facilitated the classification and report generation of their model. Visual features were extracted from the radiographic images, and then attention mechanisms and graph convolutions were applied to learn the graph embedded features. 

Liu et al \cite{LiuF2021} utilised a graph for a module in their transformer-based architecture, called the posterior-and-prior knowledge exploring-and-distilling model. They built their graph from the training splits of MIMIC-CXR and IU-Xray, where common topics are defined as nodes and are grouped by organ or body part. {In another work, Liu et al. \cite{liu2021autoencoding} again }generated their knowledge graph from the training set of the MIMIC-CXR dataset, while Zhao et al. \cite{Zhao2022} used IU-Xray and CH-Xray datasets. The majority of approaches, such as \cite{LiuF2021, liu2021autoencoding, Zhao2022, WangF2022, Li2023}, consider abnormalities as nodes while edge weights are calculated by the normalised co-occurrence of different nodes within their corpus, often a training set of an ARRG dataset.

Zhang et al. \cite{Zhang2022} constructed their knowledge graph at first by using CheXpert \cite{CheXpert} to extract and classify the given medical report for 14 common diseases, and then used Sentencepiece \cite{sentencepiece} to mine reports and extract any additional disease labels not found by CheXpert. Edges were defined by the amount of co-occurrences, with two nodes considered to be associated if they occur together more than the calculated average co-occurrenced rate. Hou et al. \cite{Hou2023} adopted CheXbert \cite{Smit2020} to help create their observation graph
which were then aligned with the visual features from the input radiographic image. Once trained, the observation graph and radiographic images are considered as a pair during inference to guide the report generation process. 

{Yang et al. \cite{Yang2022} implemented a CNN-Transformer architecture, enhanced with two forms of knowledge graphs, general and specific. The general knowledge graph was based on RadGraph \cite{Radgraph} while the specific knowledge graph was extracted from reports retrieved from their corpus of radiology reports, after which the RadGraph relation extractor was applied to extract entities and relations. The graphs were fused with the input using knowledge-enhanced multi-head attention. Wang et al. \cite{Wang2022b}'s approach, a Transformer architecture with graph-based distinctive attention, retrieved the $K$ most relevant images and used them to construct a knowledge graph. The latent representation of each image served as a node, with the $\mathcal{L}_2$ distance score between paired images  representing the weight of an edge in the graph, which was leveraged to extract more meaningful semantic features. Wang et al. \cite{WangS2022} injected prior knowledge through a knowledge graph that was constructed through text mining the RadLex Lexicon \cite{Langlotz2006}.}

{Yan introduced MaKG \cite{YanS2022}, a memory aligned knowledge graph of clinical abnormalities which sought to better learn the visual patterns of abnormalities and their relationships. Similar to Chen et al. \cite{Chen2020}, they implemented memory slots to record the features of abnormal regions from which graph embeddings were computed through the use of multi-headed attention followed by graph attentional layer. Each node embedding was used to predict the probability of an abnormality being present. Their transformer-based decoder then generated a report by attending to their graph's node embeddings. }


Shang et al. \cite{Shang2022} used a fully-connected graph to guide their multimodal encoder, with each image patch and text token treated as an independent, fully connected node. The weight of the edges are decided by an attention mechanism called optional cross attention. 
Yan et al. \cite{YanS2023} introduced a method to automatically generate a fine-grained, attributed abnormality graph, consisting of interconnected abnormality nodes and attribute nodes. Their method generated the graph through dataset annotations, radiological reports, and the RadLex lexicon \cite{Langlotz2006}.



\section{Evaluation Methods}
\label{subsec:eval}

\begin{table*}
\scriptsize
    \centering
    \begin{tabular}{lcccccccccp{13mm}c} \hline 
         \textbf{~~~~~~~~~~~~Model}  &\textbf{Year}&  \textbf{BLEU}&  \textbf{ROUGE}& \textbf{CIDEr} &\textbf{METEOR}   &\textbf{fact \cite{Miura2021}}&\textbf{AUC}&\textbf{\parbox{1cm}{\centering ~\newline Clinical \\ Efficacy}}  &\textbf{\parbox{1cm}{\centering ~\newline Case \\ Study}}&\textbf{\parbox{1cm}{\centering ~\newline Clinical \\ Validation}} &\textbf{Reproducible}\\[2.5ex] \hline 
          R2Gen \cite{Chen2020} &2020&\checkmark& \checkmark&  & \checkmark &&& \checkmark&\checkmark&  &\checkmark\\ \hline      
          Syeda-Mahmood et al. \cite{SyedaMahmood2020} &2020&\checkmark&\checkmark& & \checkmark &&&  &&  &\\ \hline 
          Zhang et al. \cite{ZhangY2020} &2020&\checkmark&\checkmark& \checkmark&   &&\checkmark&  &&  &\\ \hline 
          Lovelace \& Mortazavi \cite{Lovelace2020} &2020&\checkmark&\checkmark& \checkmark&   \checkmark &&&  \checkmark&&  &\checkmark\\ \hline 
          Boag et al. \cite{Boag2020} &2020&\checkmark&\checkmark& \checkmark&   \checkmark &&&  \checkmark&&  &\checkmark \\ \hline 
          \rowcolor[gray]{0.95}
          VTI \cite{Najdenkoska2021} &2021& \checkmark& \checkmark& & \checkmark& & \checkmark& \checkmark& & &\checkmark\\ \hline 
          \rowcolor[gray]{0.95}
          CMCL \cite{Liu2021}  &2021&\checkmark& \checkmark&  \checkmark&   &&&  &&  &\\\hline
          \rowcolor[gray]{0.95}
          CA \cite{Liu2021b}  &2021&\checkmark& \checkmark&  &   \checkmark &&&  \checkmark&\checkmark& &\checkmark \\\hline
          \rowcolor[gray]{0.95}
          Nooralahzadeh et al. \cite{Nooralahzadeh2021}  &2021&\checkmark& \checkmark&  &   \checkmark &&&  \checkmark&& &\checkmark \\\hline
          \rowcolor[gray]{0.95}
          Yan et al. \cite{Yan2021}  &2021&\checkmark&\checkmark&   &   \checkmark &&&  &&\hfil \checkmark &\checkmark \\\hline
          \rowcolor[gray]{0.95}
          AlignTransformer \cite{You2021} &2021& \checkmark& \checkmark& &   \checkmark &&&  &\checkmark&\hfil \checkmark &\\\hline
          \rowcolor[gray]{0.95}
          PPKED \cite{LiuF2021} &2021& \checkmark& \checkmark& \checkmark&   &&&  &\checkmark& &\\\hline
          \rowcolor[gray]{0.95}
          KGAE \cite{liu2021autoencoding} &2021& \checkmark& \checkmark&  &   \checkmark &&&  \checkmark&\checkmark& &\\\hline
          \rowcolor[gray]{0.95}
          R2GenCMN \cite{Chen2021} &2021& \checkmark& \checkmark&  &   \checkmark &&&  \checkmark&\checkmark& &\checkmark \\\hline
          \rowcolor[gray]{0.95}
          Nguyen et al. \cite{Nguyen2021} &2021& \checkmark& \checkmark&  &   \checkmark &&\checkmark&  \checkmark&&\hfil \checkmark &\checkmark \\\hline
          \rowcolor[gray]{0.95}
          MedWriter \cite{YangX2021} &2021& \checkmark& \checkmark&  \checkmark&   &&\checkmark&  && \hfil\checkmark &\\\hline
          \rowcolor[gray]{0.95}
          CXR-RePaiR \cite{Endo2021}  &2021& BLEU-2& &  &   &&&  \checkmark&\checkmark& &\checkmark \\\hline
          \rowcolor[gray]{0.95}
          CDGPT2 \cite{Alfarghaly2021}  &2021&\checkmark&\checkmark&  \checkmark&   \checkmark &&&  &\checkmark& \hfil \checkmark &\checkmark\\ \hline 
          \rowcolor[gray]{0.95}
          Wang et al. \cite{WangZ2021}  &2021&\checkmark& \checkmark&  \checkmark&   &&&  &\checkmark& &\\ \hline
          \rowcolor[gray]{0.95}
          Muira et al. \cite{Miura2021}  &2021&\checkmark& &  \checkmark&   &&\checkmark &  &\checkmark& &\checkmark \\ \hline
          \rowcolor[gray]{0.95}
          Medical-VLBERT \cite{LiuG2021} & 2021 &\checkmark &\checkmark &\checkmark &&&&& \checkmark &  &\\ \hline
          
          \rowcolor[gray]{0.9}
          Zhao et al. \cite{Zhao2022} &2022& \checkmark& \checkmark&  &   \checkmark &&&  &\checkmark& &\\ \hline
          \rowcolor[gray]{0.9}
          Wang et al. \cite{Wang2022}  &2022&\checkmark&\checkmark&   \checkmark&   &&&  &\checkmark& &\\\hline
          \rowcolor[gray]{0.9}
          Radgraph Reward  \cite{Delbrouck2022}  &2022&BLEU-4& \checkmark&  &   &\checkmark&&  \checkmark&\checkmark& &\\\hline       
          \rowcolor[gray]{0.9}
          Yang et al. \cite{Yang2022b} &2022&\checkmark& \checkmark&  \checkmark&   \checkmark &&\checkmark&  \checkmark&\checkmark& &\\\hline
          \rowcolor[gray]{0.9}
          Qin and Song \cite{Qin2022}  &2022&\checkmark& \checkmark&  &   \checkmark &&&  \checkmark&& &\\\hline
          \rowcolor[gray]{0.9}
          Yang et al. \cite{Yang2022}  &2022&\checkmark& \checkmark&  \checkmark&   &&&  \checkmark&& &\\\hline
          \rowcolor[gray]{0.9}
          VTI-TRS \cite{Najdenkoska2022}  &2022&\checkmark& \checkmark&  &   \checkmark &&&  &\checkmark& &\checkmark \\\hline
          \rowcolor[gray]{0.9}
          XPRONET \cite{WangJ2022}  &2022&\checkmark& \checkmark&  &   \checkmark &&&  &\checkmark& &\checkmark \\\hline
          \rowcolor[gray]{0.9}
          CMCA \cite{Song2022}  &2022&\checkmark& \checkmark&  &   \checkmark &&&  \checkmark&\checkmark& &\\\hline
          \rowcolor[gray]{0.9}
          MaKG \cite{YanS2022}  &2022&\checkmark& \checkmark&  \checkmark&   \checkmark &&\checkmark&  \checkmark&& &\\\hline
          \rowcolor[gray]{0.9}
          Zhang et al. \cite{Zhang2022} &2022&\checkmark& \checkmark&  &   \checkmark &&\checkmark&  \checkmark&\checkmark& &\\\hline
          \rowcolor[gray]{0.9}
          Nazarov et al. \cite{Nazarov2022}  &2022&\checkmark& \checkmark&  \checkmark&   \checkmark &&&  \checkmark&\checkmark& &\checkmark \\\hline
          \rowcolor[gray]{0.9}
          MATNet \cite{Shang2022} &2022&\checkmark& \checkmark&  &   \checkmark &&\checkmark&  \checkmark&& &\\\hline
          \rowcolor[gray]{0.9}
          Dalla Serra et al. \cite{DallaSerra2022} &2022&\checkmark& \checkmark&  &   \checkmark &&&  \checkmark&& \hfil \checkmark &\\\hline     
          \rowcolor[gray]{0.9}
          SGF \cite{Li2022}  &2022&\checkmark& \checkmark&  &   \checkmark &&&  &\checkmark& &\\\hline
          \rowcolor[gray]{0.9}
          JPG \cite{You2022}  &2022&\checkmark& \checkmark&  &   \checkmark &&&  && &\\\hline 
          \rowcolor[gray]{0.9}
          MedViLL \cite{Moon2022} &2022&&&&&&&&&  &\checkmark \\\hline
          \rowcolor[gray]{0.9}
          AMLMA \cite{Gajbhiye2022}  &2022&\checkmark& \checkmark&  \checkmark&   \checkmark &&&  &\checkmark& &\\\hline  
          \rowcolor[gray]{0.9}
          Relation-paraNet \cite{WangF2022} &2022&\checkmark& \checkmark& \checkmark&   &&&  &\checkmark& &\\\hline
          \rowcolor[gray]{0.9}
          Clinical-BERT \cite{YanB2022} &2022&\checkmark& \checkmark&  \checkmark&   \checkmark &&&  \checkmark&\checkmark& &\\\hline
          \rowcolor[gray]{0.9}
          RepsNet \cite{Tanwani2022} &2022&\checkmark& & &   &&&  &\checkmark& &\checkmark \\\hline
          \rowcolor[gray]{0.9}
          MSAT \cite{WangZ2022} &2022&\checkmark& \checkmark&  \checkmark&   \checkmark &&&  && & \checkmark \\\hline
          \rowcolor[gray]{0.9}
          Wang et al. \cite{WangS2022}  &2022&\checkmark& \checkmark&  \checkmark&   &\checkmark&&   \checkmark&& &\checkmark\\\hline
          \rowcolor[gray]{0.9}
          CoPlan \cite{Nishino2022} &2022& BLEU-4& \checkmark& & & & \checkmark& & & &\\\hline
          \rowcolor[gray]{0.9}
          Yan et al. \cite{Yan2022} & 2022& \checkmark& \checkmark& \checkmark& \checkmark& & &\checkmark & \checkmark& &\\\hline
          \rowcolor[gray]{0.9}
          Wang et al. \cite{Wang2022b} & 2022&\checkmark&\checkmark&&\checkmark&&&&\checkmark&  &\\\hline
          \rowcolor[gray]{0.9}
          MRCL \cite{Wu2023} & 2022 & \checkmark& \checkmark& \checkmark& \checkmark& & & & &  &\\ \hline
          \rowcolor[gray]{0.9}
          Boecking et al. \cite{Boecking2022}&2022&&&&&&&&& &\checkmark\\\hline
          \rowcolor[gray]{0.9}   
          \ps{DeltaNet} \cite{Wu2022}&\ps{2022}&\checkmark&\checkmark&\checkmark&&&&\checkmark&\checkmark& &\checkmark\\\hline
          
          \rowcolor[gray]{0.85}  
          TrMRG \cite{Mohsan2023}  &2023&\checkmark& \checkmark&  \checkmark&   \checkmark &&&  && &\\\hline
          \rowcolor[gray]{0.85}
          CheXPrune \cite{Kaur2023}  &2023&\checkmark&\checkmark&   \checkmark&   &&&  &\checkmark& &\\ \hline
          \rowcolor[gray]{0.85}
          Kim et al. \cite{Kim2023}  &2023&\checkmark& \checkmark&  \checkmark&   &&&  && &\\\hline
          \rowcolor[gray]{0.85}
          MKCL \cite{Hou2023b}  &2023&\checkmark& \checkmark&  \checkmark&   &&&  &\checkmark& &\checkmark\\\hline
          \rowcolor[gray]{0.85}
          CvT-212DistilGPT2 \cite{Nicolson2023}  &2023&\checkmark& \checkmark&  \checkmark&   \checkmark &&&  \checkmark&& &\checkmark\\\hline
          \rowcolor[gray]{0.85}
          ASGK \cite{Li2023} &2023& \checkmark&\checkmark& \checkmark&   &&&  && &\checkmark \\\hline
          \rowcolor[gray]{0.85}
          BioViL-T \cite{Bannur2023} &2023& BLEU-2&&  &   &&&  \checkmark&& &\checkmark \\\hline
          \rowcolor[gray]{0.85}
          ORGAN \cite{Hou2023}  &2023&\checkmark&\checkmark&   &   \checkmark &&&  \checkmark&\checkmark& &\\ \hline
          \rowcolor[gray]{0.85}
          X-REM \cite{Jeong2023}  &2023&BLEU-2&&   &   &&&  \checkmark&& &\checkmark \\ \hline 
          \rowcolor[gray]{0.85}
          MRCL \cite{Wu2023}  &2023&\checkmark& \checkmark&  \checkmark&   \checkmark &&&  && &\\\hline
          \rowcolor[gray]{0.85}
          KiUT \cite{Huang2023}  &2023&\checkmark& \checkmark& &   \checkmark &&&  &\checkmark& & \checkmark \\\hline
          \rowcolor[gray]{0.85}
          TIMER \cite{WuY2023}  &2023&\checkmark& \checkmark&  &   \checkmark &&&  \checkmark&\checkmark& &\checkmark \\\hline
          \rowcolor[gray]{0.85}
          Yang et al. \cite{Yang2023} &2023& \checkmark& \checkmark&  \checkmark&   &&&  \checkmark&\checkmark& &\\\hline
          \rowcolor[gray]{0.85}
          Kale et al. \cite{Kale2023} &2023& \checkmark& \checkmark&  \checkmark&   \checkmark &&&  \checkmark&& \hfil \checkmark &\\\hline
          \rowcolor[gray]{0.85}
          METransformer \cite{Wang2023} &2023& \checkmark& \checkmark&  \checkmark&   \checkmark &&&  \checkmark&\checkmark& &\\\hline
          \rowcolor[gray]{0.85}
          ATAG \cite{YanS2023} &2023&\checkmark& \checkmark&  \checkmark&   &&\checkmark&  \checkmark&& &\\\hline
          \rowcolor[gray]{0.85}
          Selivanov et al. \cite{Selivanov2023} &2023& \checkmark& \checkmark& \checkmark& & & \checkmark& \checkmark& & &\\\hline
          \rowcolor[gray]{0.85}
          RGRG \cite{tanida2023} & 2023 &\checkmark &\checkmark &\checkmark &\checkmark & & & \checkmark& \checkmark&  &\checkmark \\\hline
          \rowcolor[gray]{0.85}
          Gu et al. \cite{Gu2023} & 2023 &\checkmark&\checkmark&\checkmark&&&&&\checkmark&\hfil \checkmark &\\\hline
          \rowcolor[gray]{0.85}
          \ps{UAR} \cite{Li2023b} & \ps{2023} &\checkmark&\checkmark&\checkmark&\checkmark&&&&\checkmark&\hfil  &\\\hline
          \rowcolor[gray]{0.85}
          \ps{Dalla Serra et al.} \cite{DallaSerra2023} & \ps{2023} &\checkmark&\checkmark&&\checkmark&&&\checkmark&\checkmark&\hfil  &\\\hline
          \rowcolor[gray]{0.85}
          \ps{Zhu et al.} \cite{Zhu2023} & \ps{2023} &\checkmark&\checkmark&&\checkmark&&&\checkmark&\checkmark&\hfil  &\\\hline
          \rowcolor[gray]{0.85}
          \ps{Nguyen et al.} \cite{Nguyen2023} & \ps{2023} &\ps{BLEU-2}&&&&&&\checkmark&\checkmark&\hfil  &\\\hline          
    \end{tabular}
    \caption{Various recent methods and how they assess the performance of their model. Most ARRG researchers use NLP metrics such as BLEU, ROUGE, CIDEr and METEOR or ARRG specific methods such as Clinical Efficacy metrics. Other metrics such as fact \cite{Miura2021} and Area under the Curve are less frequently used. In terms of qualitative evaluation, a significant number of approaches perform their own case study, while some seek clinical experts to externally validate their models. Empty cells suggest an approach did not use that evaluation method.}
    \label{tbl:evaluation}
\end{table*}

The aim of ARRG is to generate a radiology report that is semantically equivalent to the report a radiologist would create. Assessing the quality of the generated report is a challenging and subjective task \cite{YuF2023}. Generated reports are required to be fluent and grammatically and  factually correct. Table~\ref{tbl:evaluation} demonstrates the metrics that are commonly adopted in the ARRG literature, 
noting many 
were originally designed for NLP tasks, such as machine translation \cite{bleu, meteor}, summarisation \cite{rouge} and image captioning \cite{cider}.

{\bf NLP Metrics --} 
The common NLP evaluation metrics used are:
\begin{itemize}
\item \textbf{BLEU - Bilingual Evaluation Understudy \cite{bleu} --}  BLEU is a precision-based metric originally designed for machine translation by analysing n-grams 
up to a length of four (hence for eaxample BLEU-1 for unigrams or BLEU-3 for trigrams etc). It primarily penalises or rewards the generated text depending on how it matches the reference text in length, word choice, and order.

\item \textbf{METEOR - Metric for Evaluation of Translation with Explicit ORdering \cite{meteor} --}  This is a precision and recall-based measure which extends BLEU-1. Originally designed for machine translation, it matches unigrams of the generated text to the reference text based on their exact form, stemmed form and meaning. It computes the harmonic mean of unigram precision and recall, biased towards recall. A multiplicative factor is used to reward identically ordered unigrams.

\item \textbf{ROUGE - Recall-Oriented Understudy for Gisting Evaluation \cite{rouge} --}  ROUGE is a set of metrics, of which ROUGE-L is employed within ARRG. It is a recall-oriented metric originally designed for summarisation tasks and measures the length of the longest common subsequence between two texts. ROGUE-L calculates the weighted harmonic mean of precision and recall, favouring recall. It does not require consecutive matches but in-sequence matches that reflect sentence level word order as n-grams. 

\item \textbf{CIDEr -- Consensus-based Image Description Evaluation  \cite{cider} --} This is also a precision and recall based metric originally developed to evaluate image captioning. CIDEr evaluates the likeness between the generated text and reference text by measuring the cosine similarity between the Term Frequency-Inverse Document Frequency weighting of n-grams, stressing the importance of content words in the evaluation. The metric also introduces a Gaussian penalty factor that rewards length similarity between the generated and reference texts.
\end{itemize}

The metrics described above focus on evaluating the lexical similarity between a generated and reference report, but there is an ongoing debate that it is more important to assign higher scores to semantically equivalent reports \cite{Boag2020, Babar2021}. For example, the sentences ``The heart is within normal size and contour.'' and  ``No observed cardiomegally.'' are semantically equivalent but would score zero on the NLP metrics commonly used. 

An interesting metric that has not been picked up in ARRG, but has been implemented for NLP tasks, such as machine translation, summarisation and image captioning, is BERTScore \cite{BERTScore}, where rather than assessing the generated sentences, token similarity is computed using contextual embeddings which enables a more robust assessment for semantically equivalent reports. Yu et al. \cite{YuF2023} quantitatively examined the correlation between automated metrics (such as BLEU and BERTScore) and the scoring of radiology reports by radiologists to understand how to meaningfully measure progress on ARRG. They found that n-gram matching methods performed worse than semantic methods such as BERTScore in evaluating the false prediction of a finding, concluding that BERTScore also had better radiologist alignment.

\subsection{Clinical Efficacy }
{A number of works have attempted to address the issue of evaluating semantic equivalence by developing clinical efficacy (CE) metrics that focus on the medical relevance and accuracy of the generated content. Chen et al. \cite{Chen2020} developed an evaluation method employing the rule-based labeller CheXpert \cite{CheXpert}, which generates 14 labels for the generated and reference reports relating to thoracic pathologies and support devices. If two reports are semantically equivalent then the derived CE score 
will be high as both reports will have received the same labels from the CheXpert labeller. Chen et al \cite{Chen2020}'s method has been utilised to assess an ARRG models performance by various researchers \cite{Chen2021, DallaSerra2022, Moon2022, Yang2022, Wang2023, Nooralahzadeh2021, YanS2023, Endo2021}.}
Endo et al. \cite{Endo2021} utilised CheXpert to create their own measure,  $S_{emb}$, to evaluate semantic equivalence through cosine similarity. It is important to note that a limitation of these CheXpert-based metrics is that they can only be utilised for models using the MIMIC-CXR dataset, as the labeling schema of CheXpert was designed with this dataset in mind.

Yan et al. \cite{YanS2023} proposed a metric called Radiology Report Quality Index (RadRQI) to evaluate the accuracy of radiology related abnormalities with the clinical attributes. Given a radiology report, keywords related to abnormalities and their attributes are extracted using RadLex \cite{Langlotz2006}. The association between abnormalities and their attributes are then contextualised by RadGraph \cite{Radgraph}, ensuring negations are considered. The proposed RadRQI-F1 score aimed to reflect the correctness of mentioned abnormality with associated attributes. In addition, the number of abnormality categories with non-zero F1 score, denoted as RadRQI-Hits, is also reported to show the coverage of distinct abnormality categories in the generated reports.

\begin{table*}[ht!]
\scriptsize
    \centering
    \begin{tabular}{llcrrrrrrr} \hline 
         \textbf{~~~~~~~~~~~~Model}   &\textbf{\parbox{2cm}{~\\Journal~or \\ Conference}}&\textbf{Year}&\textbf{BLEU-1}&  \textbf{BLEU-2}&  \textbf{BLEU-3}&  \textbf{BLEU-4}&  \textbf{ROUGE}& \textbf{CIDEr} &\textbf{METEOR}\\[2.5ex] \hline 

          Lovelace \& Mortazavi \cite{Lovelace2020}&EMNLP&2020&0.415&  0.272&  0.193&  0.146& 0.318& 0.316&0.159\\ \hline 
          R2Gen \cite{Chen2020} &EMNLP&2020&0.470&  0.304&  0.219&  0.165& 0.371&&0.187\\ \hline 
          {Syeda-Mahmood et al. \cite{SyedaMahmood2020}} & MICCAI&2020&\underline{0.560}&  \textbf{0.510}&  \textbf{0.500}&  \textbf{0.490}&  \textbf{0.580}&&\textbf{0.580}\\ \hline 
          
          \rowcolor[gray]{0.95}
          VTI \cite{Najdenkoska2021}  &MICCAI&2021&0.493&  0.360&  0.291&  0.154&  0.375& &0.281\\ \hline
          \rowcolor[gray]{0.95}
          CDGPT2 \cite{Alfarghaly2021}  &IMU&2021&0.387&  0.245&  0.166&  0.111&  0.289&  0.257&0.164\\ \hline 
          \rowcolor[gray]{0.95}
          CMCL \cite{Liu2021}  &IJCNLP&2021& 0.473& 0.305& 0.217& 0.162& 0.378& &0.186\\\hline
          \rowcolor[gray]{0.95}
          Nooralahzadeh et al. \cite{Nooralahzadeh2021}  &EMNLP&2021& 0.486& 0.317& 0.232& 0.173& 0.390& 
          &0.192\\\hline 
          \rowcolor[gray]{0.95}
          Wang et al. \cite{WangZ2021}  & CVPR&2021& 0.487& 0.346& 0.270& 0.208& 0.359& 0.452&\\\hline
          \rowcolor[gray]{0.95}
          CA \cite{Liu2021b}  & IJCNLP&2021& 0.492& 0.314& 0.222& 0.169& 0.381& &0.193\\\hline
          \rowcolor[gray]{0.95}
          AlignTransformer \cite{You2021} & MICCAI& 2021& 0.484& 0.313& 0.225& 0.173& 0.379& &0.204\\\hline
          \rowcolor[gray]{0.95}
          PPKED \cite{LiuF2021} & CVPR& 2021& 0.483& 0.315& 0.224& 0.168& 0.376&0.351 &\\\hline
          \rowcolor[gray]{0.95}
          KGAE \cite{liu2021autoencoding} &NeurIPS& 2021& 0.512& 0.327& 0.240& 0.179& 0.383& &0.195\\\hline
          \rowcolor[gray]{0.95}
          R2GenCMN \cite{Chen2021} & IJCNLP& 2021& 0.475& 0.309& 0.222& 0.170& 0.375& &0.191\\\hline
          \rowcolor[gray]{0.95}
          Nguyen et al. \cite{Nguyen2021} &EMNLP& 2021& 0.515& 0.387& 0.293& 0.235& 0.436& &0.219\\\hline
          \rowcolor[gray]{0.95}
          MedWriter \cite{YangX2021} & IJCNLP& 2021& 0.471& 0.336& 0.238& 0.166& 0.382& 0.345&\\\hline
          \rowcolor[gray]{0.95}
          Zhao et al. \cite{Zhao2022} & ICCPR& 2021& 0.502& 0.326& 0.232& 0.176& 0.381&&0.199\\\hline
          
          \rowcolor[gray]{0.9}
          Wang et al. \cite{WangS2022}  & AMIA&2022&0.450&  0.301&  0.213&  0.158&  0.384&  0.340&\\ \hline
          \rowcolor[gray]{0.9}
          Radgraph Reward \cite{Delbrouck2022}  & EMNLP& 2022& & & & 0.139& 0.327& &\\\hline
          \rowcolor[gray]{0.9}
          Yang et al. \cite{Yang2022b}  & Neurocomputing&2022& 0.491& 0.334& 0.242& 0.18& 0.397& 0.469&0.212\\\hline
          \rowcolor[gray]{0.9}
          Qin and Song \cite{Qin2022}  & ACL&2022& 0.494& 0.321& 0.235& 0.109& 0.384& &0.201\\\hline
          \rowcolor[gray]{0.9}
          Yang et al. \cite{Yang2022}  & Med Image Anal.&2022& 0.496& 0.327& 0.238& 0.178& 0.381& 0.382& \\\hline
          \rowcolor[gray]{0.9}
          VTI-TRS \cite{Najdenkoska2022}  & Med Image Anal.&2022& 0.503& 0.394& 0.302& 0.170& 0.390& &0.230\\\hline
          \rowcolor[gray]{0.9}
          SGF \cite{Li2022}  & MICCAI&2022& 0.467& 0.334& 0.261& 0.215& 0.415& &0.201\\\hline
          \rowcolor[gray]{0.9}
          XPRONET \cite{WangJ2022}  & ECCV&2022& 0.525& 0.357& 0.262& 0.199& 0.411& 0.359&0.22\\\hline
          \rowcolor[gray]{0.9}
          JPG \cite{You2022}  & ICCL& 2022& 0.479& 0.319& 0.222& 0.174& 0.377& &0.193\\\hline
          \rowcolor[gray]{0.9}
          CMCA \cite{Song2022}  & ICCL &2022& 0.497& 0.349& 0.268& 0.215& 0.392& &0.209\\\hline
          \rowcolor[gray]{0.9}
          MaKG \cite{YanS2022}  & JBHI&2022& & & & & 0.353& 0.681&\underline{0.392}\\\hline
          \rowcolor[gray]{0.9}
          AMLMA \cite{Gajbhiye2022}  & CMPB&2022& 0.471& 0.315& 0.231& 0.172& 0.376& 0.381&0.247\\\hline
          \rowcolor[gray]{0.9}
          MaKG \cite{YanS2022}  & BioNLP&2022& & & & & 0.353& 0.523& \underline{0.378}\\\hline
          \rowcolor[gray]{0.9}
          Zhang et al. \cite{Zhang2022} & Appl. Sci.&2022& 0.505& 0.379& 0.303& 0.251& 0.446& &0.218\\\hline
          \rowcolor[gray]{0.9}
          ATAG \cite{YanS2023} &  IEEE TMI.&2022& & & & & 0.341& 0.380&\\\hline
          \rowcolor[gray]{0.9}
          Relation-paraNet \cite{WangF2022} & IEEE TC&2022& 0.505& 0.329& 0.230& 0.168& 0.372& 0.317&\\\hline
          \rowcolor[gray]{0.9}
          Clinical-BERT \cite{YanB2022} & AAAI&2022& 0.495& 0.330& 0.231& 0.170& 0.376& 0.432&\\\hline
          \rowcolor[gray]{0.9}
          MATNet \cite{Shang2022} & IEEE SPL&2022& 0.518& 0.387& 0.308& 0.254& 0.446&&0.222\\\hline
          \rowcolor[gray]{0.9}
          RepsNet \cite{Tanwani2022} &MICCAI &2022& \textbf{0.580}& 0.440& 0.320& 0.270&&&\\\hline
          \rowcolor[gray]{0.9}
          \ps{DeltaNet} \cite{Wu2022} &\ps{ICCL} &\ps{2022}& \ps{0.485}&\ps{0.324}&\ps{0.238}& \ps{0.184}&\ps{0.379}&\textbf{\ps{0.802}}&\\\hline
          
          \rowcolor[gray]{0.85}
          TrMRG \cite{Mohsan2023}  & IEEE Access&2023& 0.532& 0.344& 0.233& 0.158& 0.387& 0.500&0.218\\\hline
          \rowcolor[gray]{0.85}
          KiUT \cite{Huang2023}  & CVPR&2023& 0.525& 0.36& 0.251& 0.185& 0.409& &0.242\\\hline
          \rowcolor[gray]{0.85}
          {CheXPrune} \cite{Kaur2023}  &  JAIHC&2023&0.543&  \underline{0.445}&  \underline{0.374}&  \underline{0.320}&  \textbf{0.598}&  0.322&\\ \hline
          \rowcolor[gray]{0.85}
          MRCL \cite{Wu2023}  & JAIHC&2023& 0.458& 0.324& 0.238& 0.180& 0.369& 0.287&0.206\\\hline
          \rowcolor[gray]{0.85}
          Kim et al. \cite{Kim2023}  & BioNLP&2023& 0.438& 0.28& 0.201& 0.155& 0.351& 0.631&\\\hline
          \rowcolor[gray]{0.85}
          ORGAN \cite{Hou2023}  & JBI&2023& 0.510& 0.346& 0.255& 0.195& 0.399& &0.205\\\hline
          \rowcolor[gray]{0.85}
          MKCL \cite{Hou2023b}  & JBI&2023& 0.490& 0.311& 0.222& 0.167& 0.385& 0.523&0.206 \\\hline
          \rowcolor[gray]{0.85}
          TIMER \cite{WuY2023}  & JMLR&2023& 0.493& 0.325& 0.238& 0.186& 0.383& &0.204\\\hline
          \rowcolor[gray]{0.85}
          CvT-212DistilGPT2 \cite{Nicolson2023}  & AIMI&2023& 0.473& 0.304& 0.224& 0.175& 0.376& 
          \underline{0.694}&0.200\\\hline
          \rowcolor[gray]{0.85}
          Yang et al. \cite{Yang2023} & Med Image Anal.& 2023& 0.497& 0.319& 0.230& 0.174& 0.399& 0.407&\\\hline
          \rowcolor[gray]{0.85}
          ASGK \cite{Li2023} & WWWJ& 2023&&&& 0.125& 0.279& 0.306&\\\hline
          \rowcolor[gray]{0.85}
          \rowcolor[gray]{0.85}
          METransformer \cite{Wang2023} & CVPR& 2023& 0.483& 0.322& 0.228& 0.172& 0.380& 0.435& 0.192\\\hline
          \rowcolor[gray]{0.85}
          Gu et al. \cite{Gu2023} & Bioengineering& 2023&0.460&0.294&0.207&0.152&0.385&0.409&\\\hline
          \rowcolor[gray]{0.85}
          \ps{UAR} \cite{Li2023b} & \ps{JMLR}& \ps{2023}&\ps{0.530}&\ps{0.365}&\ps{0.263}&\ps{0.200}&\ps{0.405}&\ps{0.501}&\ps{0.218}\\\hline
    \end{tabular}
    \caption{IU-Xray Dataset: Comparative Model Performance. All values were extracted from the original papers. Best results are in bold and the second-best underlined. Empty cells denote metrics not evaluated by the model. The table is grouped by year.}
    \label{tbl:iuxray}
\end{table*}

\begin{table*}[t]
\scriptsize
    \centering

   \begin{tabular}{llcrrrrrrr} \hline 
         \textbf{~~~~~~~~~Model}   &\textbf{\parbox{2cm}{~\\Journal~or \\ Conference}}&\textbf{Year}&\textbf{BLEU-1}&  \textbf{BLEU-2}&  \textbf{BLEU-3}&  \textbf{BLEU-4}&  \textbf{ROUGE}& \textbf{CIDEr} &\textbf{METEOR}\\[2.5ex] \hline 
         
          Lovelace \& Mortazavi \cite{Lovelace2020} & EMNLP&2020&0.415&  0.272&  0.193&  0.146& 0.318& 0.316&0.159\\\hline 
          Boag et al. \cite{Boag2020} & NeurIPS&2020&0.305&  0.201&  0.137&  0.092&& \textbf{0.850}&\\ \hline 
          R2Gen \cite{Chen2020} & EMNLP&2020&0.470&  0.304&  0.219&  0.165& 0.371&&0.187\\ \hline 
          \rowcolor[gray]{0.95}
          VTI \cite{Najdenkoska2021}  &MICCAI&2021&0.418&  0.293&  0.152&  0.109&  0.302& &0.177\\ \hline 
          \rowcolor[gray]{0.95}
          CMCL \cite{Liu2021}  & IJCNLP&2021& 0.344& 0.217& 0.14& 0.097& 0.281& &0.133\\\hline
          \rowcolor[gray]{0.95}
          CA \cite{Liu2021b}  & IJCNLP&2021& 0.350& 0.219& 0.152& 0.109& 0.283& &0.151\\\hline
          \rowcolor[gray]{0.95}
          Nooralahzadeh et al. \cite{Nooralahzadeh2021}  & EMNLP&2021& 0.378& 0.232& 0.154& 0.107& 0.272& &0.145\\\hline
          \rowcolor[gray]{0.95}
          Yan et al. \cite{Yan2021}  & EMNLP& 2021&0.373&  &  &  0.107&  0.274& &0.144\\\hline
          \rowcolor[gray]{0.95}
          AlignTransformer \cite{You2021} & MICCAI& 2021& 0.378& 0.235& 0.156& 0.112& 0.283& &0.158\\\hline
          \rowcolor[gray]{0.95}
          PPKED \cite{LiuF2021} & CVPR& 2021& 0.360& 0.224& 0.149& 0.106& 0.284& &0.149\\\hline
          \rowcolor[gray]{0.95}
          KGAE \cite{liu2021autoencoding} & NeurIPS& 2021& 0.369& 0.231& 0.156& 0.118& 0.295& &0.153\\\hline
          \rowcolor[gray]{0.95}
          R2GenCMN \cite{Chen2021} & EMNLP& 2021& 0.353& 0.218& 0.148& 0.106& 0.278& &0.142\\\hline
          \rowcolor[gray]{0.95}
          Nguyen et al. \cite{Nguyen2021} & EMNLP& 2021& \underline{0.495}& 0.360& 0.278& 0.224& 0.222& &0.219\\\hline
          \rowcolor[gray]{0.95}
          MedWriter \cite{YangX2021} & IJCNLP& 2021& 0.438& 0.297& 0.216& 0.164& 0.332& 0.306&\\\hline
          \rowcolor[gray]{0.95}
          CXR-RePaiR \cite{Endo2021}  & JMLR& 2021& 0.438& 0.297& 0.216& 0.164& 0.332& 0.306&\\\hline

          \rowcolor[gray]{0.9}
          Wang et al. \cite{Wang2022}  & IEEE TMI &2022&0.373&  0.235&  0.162&  0.120&  0.282& 0.299&0.143\\\hline 
          \rowcolor[gray]{0.9}
          Radgraph Reward  \cite{Delbrouck2022}  & EMNLP&2022& & & 0.259& 0.116& & &\\\hline
          \rowcolor[gray]{0.9}
          Yang et al. \cite{Yang2022b}  & Neurocomputing&2022& 0.362& 0.224& 0.15& 0.108& 0.276& 0.157&0.142\\\hline
          \rowcolor[gray]{0.9}
          Qin and Song \cite{Qin2022}  & ACL&2022& 0.381& 0.232& 0.155& 0.109& 0.287& &0.151\\\hline
          \rowcolor[gray]{0.9}
          Yang et al. \cite{Yang2022}  & Med Image Anal.&2022& 0.363& 0.228& 0.156& 0.115& 0.284& 0.203&\\\hline
          \rowcolor[gray]{0.9}
          VTI-TRS \cite{Najdenkoska2022}  & Med Image Anal.&2022& 0.475& 0.314& 0.196& 0.136& 0.315& &0.191\\\hline
          \rowcolor[gray]{0.9}
          XPRONET \cite{WangJ2022}  & ECCV&2022& 0.344& 0.215& 0.146& 0.105& 0.279& &0.138\\\hline
          \rowcolor[gray]{0.9}
          CMCA \cite{Song2022}  & COLING&2022& 0.360& 0.227& 0.156& 0.117& 0.287& &0.148\\\hline
          \rowcolor[gray]{0.9}
          MaKG \cite{YanS2022}  & BioNLP&2022& & & & & 0.228& 0.120&\underline{0.284}\\\hline
          \rowcolor[gray]{0.9}
          Zhang et al. \cite{Zhang2022} & Appl. Sci.&2022& 0.491& 0.358& 0.278& 0.225& 0.389& &0.218\\\hline
          \rowcolor[gray]{0.9}
          ATAG \cite{YanS2023} & IEEE TMI.&2022& & & & & 0.225& 0.160&\\\hline
          \rowcolor[gray]{0.9}
          Nazarov et al. \cite{Nazarov2022}  & BMVC&2022& 0.299& 0.182& 0.124& 0.090& 0.238& 0.136&0.123\\\hline
          \rowcolor[gray]{0.9}
          {MATNet \cite{Shang2022}} & IEEE SPL&2022& \textbf{0.506}& \textbf{0.370}& \underline{0.288}& \underline{0.233}& 0.395&&0.221\\\hline
          \rowcolor[gray]{0.9}
          Dalla Serra et al. \cite{DallaSerra2022} & IJCNLP&2022& 0.363& 0.245& 0.178& 0.161& 0.313&&\\\hline
          \rowcolor[gray]{0.9}
          MSAT \cite{WangZ2022} & MICCAI&2022& 0.373& 0.235& 0.162& 0.120& 0.282& 0.299& 0.143\\\hline
          \rowcolor[gray]{0.9}
          \ps{DeltaNet} \cite{Wu2022} &\ps{ICCL} &\ps{2022}& \ps{0.361}&\ps{0.225}& \ps{0.154}& \ps{0.114}&\ps{0.277}&\ps{0.281}&\\\hline

          \rowcolor[gray]{0.85}
          Kim et al. \cite{Kim2023}  & BioNLP&2023& 0.342& 0.222& 0.152& 0.110& 0.301& 0.166&\\\hline
          \rowcolor[gray]{0.85}
          ORGAN \cite{Hou2023}  & J. Biomed. Inform.&2023&0.407&  0.256&  0.172&  0.123&  0.293&  &0.162\\ \hline
          \rowcolor[gray]{0.85}
          X-REM \cite{Jeong2023}  & MIDL&2023&&  0.22&  &  &  &  &\\ \hline 
          \rowcolor[gray]{0.85}
          MRCL \cite{Wu2023}  & JAIHC& 2023& 0.340& 0.212& 0.145& 0.103& 0.270& 0.109&0.137\\\hline
          \rowcolor[gray]{0.85}
          KiUT \cite{Huang2023}  & CVPR&2023& 0.393& 0.243& 0.159& 0.113& 0.285& &0.16\\\hline
          \rowcolor[gray]{0.85}
          TIMER \cite{WuY2023}  & JMLR&2023& 0.383& 0.225& 0.146& 0.104& 0.28& &0.147\\\hline
          \rowcolor[gray]{0.85}
          CvT-212DistilGPT2 \cite{Nicolson2023}  & AIMI&2023& 0.393& 0.248& 0.171& 0.127& 0.286& 0.389&0.155\\\hline
          \rowcolor[gray]{0.85}
          Yang et al. \cite{Yang2023} & Med Image Anal.& 2023& 0.386& 0.237& 0.157& 0.111& 0.274& 0.111&\\\hline
          \rowcolor[gray]{0.85}
          METransformer \cite{Wang2023} & CVPR& 2023& 0.386& 0.250& 0.169& 0.124& 0.291& 0.362& \textbf{0.362}\\\hline
          \rowcolor[gray]{0.85}
          RGRG \cite{tanida2023}&CVPR & 2023 & 0.373&0.249 & 0.175& 0.126&0.264 &\underline{0.495} &0.168 \\\hline 
          \rowcolor[gray]{0.85}
          \ps{UAR} \cite{Li2023b}&\ps{JMLR} & \ps{2023} & \ps{0.363}&\ps{0.229} & \ps{0.158}& \ps{0.107}&\ps{0.289} &\ps{0.246} &\ps{0.157}
          \\\hline 
          \rowcolor[gray]{0.85}
          \ps{Dalla Serra et al.} \cite{DallaSerra2023}&\ps{EMNLP} & \ps{2023} & \ps{0.486}&\underline{\ps{0.366}} & \textbf{\ps{0.295}}& \textbf{\ps{0.246}}&\textbf{\ps{0.423}} & &\ps{0.216} \\\hline 
          \rowcolor[gray]{0.85}
          \ps{Zhu et al.} \cite{Zhu2023}&\ps{MICCAI} & \ps{2023} & \ps{0.343}&\ps{0.210} & \ps{0.140}& \ps{0.099}&\ps{0.271} & &\ps{0.137} \\\hline 
          \rowcolor[gray]{0.85}
          \ps{Nguyen et al.} \cite{Nguyen2023}&\ps{JMLR} & \ps{2023} &&\ps{0.137} &&&&& \\\hline 
    \end{tabular}
    \caption{MIMIC-CXR Dataset: Comparative Model Performance. All values were extracted from the original papers. Best results are in bold and the second-best underlined. Empty cells denote metrics not evaluated by the model.}
    \label{tbl:mimic}
\end{table*}

Miura et al. \cite{Miura2021} developed two factually oriented metrics called $fact_{ENT}$ and $fact_{ENTNLI}$ which evaluate the factual correctness and completeness of the generated findings section of a report, for a more comprehensive explanation of this work please see Section~\ref{sec:reinforcement}. While Miura et al. \cite{Miura2021} primarily used these metrics as rewards to train their model using reinforcement learning, they have been applied by some during evaluation \cite{Miura2021, Delbrouck2022}. 

\subsection{Diversity Metrics}
The process of writing radiology reports is inherently ambiguous, stemming from the diverse levels of expertise, experience, and expressive styles among radiologists \cite{Najdenkoska2022}. To address this, certain ARRG researchers have endeavored to quantify these differences by developing metrics that gauge the distinctiveness of a generated report in comparison to others. Gajbhiye et al. \cite{Gajbhiye2022} introduced the metric Unique Index to demonstrate a model's ability to generate unique reports. This is calculated by taking the fraction of the total number of a unique reports a model generates against the total number of unique reports in the test set. For example, if a model created 100 unique reports from a test set of 500 unique reports, then its score would be $0.2$. Similarly, Najdenkoska et al. \cite{Najdenkoska2022} used evaluation protocols from image captioning for measuring diversity within the generated captions. The \%Novel metric they adopt measured the percentage of generated sentences that are not contained within the model's training set.

\subsection{Qualitative Evaluation}
Adequately evaluating the performance of medical texts can be challenging due to the complexity of the human language \cite{abacha2023}. To combat this, some researchers have evaluated their models in a qualitative fashion. Christy et al. \cite{Christy2018, Christy2019} performed  evaluations via Amazon Mechanical Turk by asking participants to choose amongst reports generated by different models that match with the ground truth report in terms of language fluency, content selection, and correctness of abnormal medical findings. More recent research has utilised clinical experts to evaluate their models \cite{Yan2021, liu2021autoencoding, You2021, YangX2021, Alfarghaly2021, DallaSerra2022, Kale2023}.
For example, Li et al. \cite{Li2023} engaged three radiologists to evaluate the outputs of their methods, which is explained in Section~\ref{subsec:training}, against CoAtt and Vision-BERT.

Although using human clinical experts to evaluate medical machine learning models is considered the most reliable method \cite{abacha2023}, it is time consuming and costly \cite{YuF2023}. Asking healthcare professionals to evaluate the generated reports provides more work for an already overstretched workforce. Finding methods to accurately evaluate not only the fluency of the generated reports, but also the medical semantics are an important avenue of research.
        
\section{Performance Comparison}
\label{sec:evaluation}
In this section, we scrutinise the results of the state of the art in ARRG by comparing their reported metrics, including BLEU, ROUGE, and METEOR. All values referred to have been extracted from the original papers, and where the information was missing, it was assumed the same training and test set from the respective dataset (IU-Xray and MIMIC-CXR) were used for each.

For the IU-Xray dataset, Table~\ref{tbl:iuxray} illustrates that the most effective model according to several metrics is by Syeda-Mahmood et al. \cite{SyedaMahmood2020}. Their model extracted not just core findings relating to pathologies, but also findings relating to location, size and severity which they coined fine-finding labels. \ps{These labels were used to retrieve the most similar report from their report corpus, after which the report was edited to more closely align with the generated labels. Regrettably, their code has not been made public, posing a challenge for exact reproducibility and their evaluation does not contain an ablation study which would have helped quantify the effectiveness of their contribution. Their paper did provide a small case study which compared their model's generated reports against the ground truth and a select few other models which highlighted their reports' succinct nature, possibly due to their report editing process. CheXPrune by Kaur \& Mittal \cite{Kaur2023} also performed well, achieving the second highest score on most NLP metrics. Their method adopted a multi-attention mechanism and utilised parameter pruning to create a more lightweight model. Their approach} applied a hierarchical-LSTM for decoding which has previously been accused of generating inflexible reports \cite{Wang2022, Gajbhiye2022}. Unfortunately, Kaur \& Mittal \cite{Kaur2023} have not released their code for reproducibility \ps{and their own evaluations did not provide any ablation studies or other case studies, hindering an in-depth evaluation of their work. Despite this, their performance implies that the LSTM architecture, coupled with appropriate supporting modules, can generate reports on a par, and above, the more recent architectures.} 

RepsNet \cite{Tanwani2022} also performed strongly, leveraging publicly available models to obtain the highest BLEU-1 score. They used ResNeXt \cite{ResNeXt} and BERT \cite{BERT} as their encoders and GPT-2 \cite{Radford2019} as their decoder to generate their final reports, training with contrastive loss. \ps{During their discussion, Tanwani et al. \cite{Tanwani2022} evaluated the effectiveness of their model through the use of ablation and qualitative case studies. Their ablation study presented results for a base model before incrementally adding two techniques, contrastive learning and inserting prior context from retrieved reports. They observed that their model performed best when using their complete methodology, demonstrating that LLMs based methods such as RepsNet} are capable of producing state of the art results within the ARRG domain.

Table~\ref{tbl:mimic} demonstrates performances on the MIMIC-CXR dataset with the best scoring model being produced \ps{by Dalla Serra et al. \cite{DallaSerra2023}, whose contribution leveraged images from previous scans and clinical indications to improve their model's performance. While there is also no code released for this work, they performed an indepth performance analysis. Their ablation study investigated the effectiveness of incorporating prior scans as input and adopting sentence-anatomy (SA) dropout during training, where a random subset of anatomical tokens was dropped from the input and omitted from the generated radiology report. The findings from this investigation demonstrated  a marked improvement in performance from the baseline with the addition of either technique. There was also a further improvement  when their model implemented both techniques which demonstrates the effectiveness of both their training and multimodal input strategies. They also investigated the effectiveness of their model with regard to initial and follow up scans by splitting the test set respectively. This study highlighted that including priors improved their model's performance on not only follow up examinations, but also on imaging studies with no prior scans. This improvement was reasoned to occur because their model could infer which samples are initial scans from the zero filled vectors that were used as a placeholder instead of an image.} MATNet by Shang et al. \cite{Shang2022} also performed well, scoring second in five metrics. \ps{Similarily to Dalla Serra et al. \cite{DallaSerra2023}, their approach also tried to incorporate extra knowledge into their model in the form of the clinical indications provided with the MIMIC-CXR dataset as a secondary input to their encoder. Unfortunately, Shang et al. \cite{Shang2022} did not publically release their code which makes difficult to replicate their results but their ablation study demonstrated the effectiveness of the individual components of their model. They provided results for a base model on the IU-xray dataset and then added components, including their multi-modal Encoder, disease classifier, and adaptive decoder. The performance of their model drastically improved with the addition of all three components, with the best performance being achieved when all three additional components are used.}

\ps{Reflecting on the results, it is vital that the reporting of dataset usage within research is transparent and standardised especially in terms of the data split. While the MIMIC-CXR dataset provided its own, IU-Xray does not contain an official split, and our recommendation for IU-Xray is to follow the split provided by Chen et al \cite{Chen2020} to facilitate accurate comparison of results across different research.} \ps{Furthermore, when examining Table~\ref{tbl:iuxray} and Table~\ref{tbl:mimic}, it becomes evident that there is considerable variation in the NLP metrics employed, often hindering accurate model comparisons. To ensure reliable comparisons, it is advisable to adopt the NLP metrics BLEU, ROUGE-L, CIDEr, and METEOR as a minimum benchmark.}

\ps{Among the reviewed models, it is important to note that IU-Xray is a much smaller dataset than MIMIC-CXR and as such, IU-Xray's variance in results could be significant \cite{Chen2021}. As MIMIC is larger, and has an official split which can be used to standardise results, it is considered to be a more accurate baseline for the current state of the art within the field. On the MIMIC-CXR dataset, the two highest performing models for the MIMIC dataset both utilised multimodal inputs to improve their scores which had been validated by their own ablation and case studies within their papers. Due to this, it is reasoned that adding extra information, either through temporal or multimodal approaches, can be considered the current state of the art approach within ARRG.}

\section{Conclusion}
\label{sec:conclusion}
ARRG poses an inherently intricate challenge for deep learning, addressing complexities from both computer vision and natural language generation. In this review, we have conducted a methodological assessment of deep learning research methods addressing the ARRG task. We offer a description of the ARRG task, followed by an overview of the datasets commonly employed in the current literature. The majority of approaches utilised the two main datasets, IU-Xray and MIMIC-CXR, with a minority exploring datasets that are not publicly available. For the review, we analysed 70 ARRG contributions which were published between 2020 to 2023, categorising their contributions into specific topic areas, including training methods, architecture, utilising knowledge and multiple modalities and evaluation methods.

Given the lack of a universally accepted standard for evaluation, we are unable to advocate for an ideal model architecture, however from the papers reviewed we can infer some recommendations. Methods have highlighted the benefits of utilising auxiliary tasks, either in the pre-training phase or throughout the training process. In terms of architecture, by observing Table~\ref{tbl:architectures} we can see that the use of the transformer is growing in popularity. This trend is expected to continue, with foundational models such as LLaMA \cite{llama} being adopted within the domain. Exploring ways to integrate knowledge into ARRG models has emerged as a central focus for numerous studies within the field.

Methods for evaluating ARRG systems is still an open research question. Commonly used NLP metrics are not domain specific and cannot be utilised to measure the diagnostic quality of generated radiology reports \cite{Boag2020}. Clinical efficacy metrics \cite{Chen2020} are a step in the right direction, but these measures depend on specific methods used to construct the rule-based classifier, meaning the metric is specific to a single dataset \cite{Babar2021}. The design of metrics that can better handle semantics and assess clinical evaluations is essential.
Our evaluation section provides insights into the performance of the designed models and showcases their results. We found that despite encoder-decoder models being the more prevalent architecture, retrieval based approaches perform at a similar level on NLP metrics. 
Another insight is that the introduction of multimodal inputs have improved performance.

\subsection{Future Directions}
Finally, we highlight unresolved issues in the literature that create opportunities for potential future research directions within the ARRG field.

\textbf{Better Utilisation of Clinical Experts --} If the ultimate objective of ARRG is to create a radiology report generation system, it must adhere to the same rigorous standards expected of radiologists and other reporting clinical experts. As these experts are required to be qualified and experienced in order to report, it makes sense to take advantage of their abilities within the ARRG workflow where possible. A minority of approaches already incorporated reviews by clinical experts into their studies \cite{You2021, Yan2021, Alfarghaly2021, YangX2021, Nguyen2021, DallaSerra2022, Kale2023} but providing an objectively fair comparison between the approaches is challenging due to the lack of standardisation of these reviews. Consequently, establishing a standardised approach to qualitative evaluation would facilitate a more straightforward comparison of approaches that adopt these assessments by clinical experts. 

{Reinforcement learning is becoming more prevalent in ARRG, however an area that is currently unexplored is reinforcement learning with human feedback \cite{Christiano2017} which has been successfully used to improve the performance of LLMs \cite{ChatGPT}. Learning a reward function from the feedback data of clinical experts is a promising avenue for ARRG as their feedback can promote considerations such as safety and ethics which are big areas of concern within the medical field, while also reducing  overall model training costs.} 

\textbf{Improving Quantitative Evaluation --} While qualitative evaluation from a clinical expert is beneficial, it is both costly and impractical for these experts to assess all cases. Having an agreed set of quantitative metrics that evaluate the semantics, quality and correctness of generated reports is crucial. As seen in Table~\ref{tbl:evaluation}, the most commonly adopted assessment criteria are NLP metrics which have been found to fall short for ARRG assessment \cite{Boag2020}.


Several clinical metrics have been proposed, but a universally accepted gold standard metric has not yet been established. Consequently, the challenge of developing metrics to assess medical accuracy and correctness remains an ongoing research challenge. Metrics that have been used within machine translation and summarisation could be used within the ARRG domain to compare generated and reference reports. Metrics such as BERTScore \cite{BERTScore}, which use embeddings to estimate semantic distance rather than relying on exact matches between n-grams, have been found to be more effective than metrics such as ROUGE, BLEU and METEOR.

\textbf{Application of Large Language Models --} While examining current research, it was observed that there exists a gap in the literature concerning the application of \ps{state-of-the-art} large language models. \ps{These models have been used and evaluated for similar tasks such as simplifying radiology reports or generating impressions from findings \cite{Sun2023, Amin2023}}. However, the adoption of LLM decoder-only models within ARRG research has been restricted to models from previous generations, such as GPT-2 \cite{Alfarghaly2021} or the utilisation of the ChatGPT web interface \cite{Chaoyi2023}. Foundational models such as LLaMA \cite{llama} and Falcon \cite{falcon} have been released in multiple versions, with LLaMA ranging from 7 to 65 billion parameters. These open-source models offer greater accessibility to the research community and have not yet been incorporated into the ARRG task.  Given their demonstrated state-of-the-art performance in numerous NLP tasks, it is expected that the incorporation of these models into research is imminent.

\textbf{Adoption of new datasets --} The ARRG research community has focused their efforts on two datasets: IU-Xray and MIMIC-CXR. Both datasets share the same imaging modality, chest X-rays. Given the availability of these datasets, the community's emphasis \ps{has been} generating reports specifically for chest X-rays. As a resource, PadChest remains largely untapped, and incorporating it into research could enhance the ability of future models to generalise. \ps{Hamamci et al \cite{Hamamci2024} have recently compiled a new public 3D dataset from an alternative imaging modality, CT. This dataset offers significant value to the ARRG community, affording them the opportunity to broaden the domain's scope and explore new avenues of research.} 

\ps{\textbf{Extending model capabilities --} The bulk of the examined research has been focused on generating the findings of a radiology report, with only a limited number looking into generating the impressions section. Extending models to generate the impression section is an interesting area to explore and could be approached from various perspectives. For instance, this could involve concatenation of the findings and impression sections to be used as a target in generation, or using NLP summarization techniques on the generated findings to formulate an impression section.}

\section*{Data access statement}
No new data was created during this review article.

\section*{Acknowledgment}
The 1st Author wishes to acknowledge and thank the financial support of the UKRI (Grant ref EP/S022937/1) and the University of Bristol. 

\end{document}